%% file: ml_cmf.tex
\documentclass[a4paper,10pt]{article}
\usepackage[left=3cm,right=3cm]{geometry}
\usepackage[T1]{fontenc}
\usepackage[utf8]{inputenc}
\usepackage[USenglish]{babel}
\usepackage{amsthm}
\usepackage{amsmath}
\usepackage{amssymb}
\usepackage{braket,amsfonts}
\usepackage{bm} % Used for boldface symbol
\usepackage{accents}
\usepackage{array}

\newtheorem{lemma}{Lemma}
\newtheorem{theorem}{Theorem}
\newtheorem{remark}{Remark}
\usepackage{algpseudocode}
\usepackage{algorithm}

\usepackage{authblk}

\usepackage{url} % for url in the references
\usepackage{multirow}
\usepackage{setspace}
\usepackage{subcaption}

\usepackage{pgfplots}

\usepackage{amsopn}

\usepackage{xspace}
\usepackage{bold-extra}
\usepackage[most]{tcolorbox}

\colorlet{texcscolor}{blue!50!black}
\colorlet{texemcolor}{red!70!black}
\colorlet{texpreamble}{red!70!black}
\colorlet{codebackground}{black!25!white!25}
\patchcmd\newpage{\vfil}{}{}{}
\title{Machine learning-based conditional mean filter: \\
a generalization of the ensemble Kalman filter for nonlinear data assimilation}

\author[1]{Truong-Vinh Hoang} 
\author[1]{Sebastian Krumscheid}
\author[2]{Hermann G. Matthies}
\author[3]{Ra\'ul~Tempone}

\affil[1]{Chair of Mathematics for Uncertainty Quantification, RWTH Aachen University, Germany\\
	hoang@uq.rwth-aachen.de}
\affil[2]{Technische Universität Braunschweig, Germany}
\affil[3]{Computer, Electrical and  Mathematical Sciences and Engineering, KAUST,
and  Alexander von Humboldt professor in Mathematics of Uncertainty Quantification, RWTH Aachen University.}
\begin{document}
	
\maketitle

%%%%%%%%%%%%%%%%%%%%%%%%%%%%%%%%%%%%%%%%%%%%%%%%%%
\begin{abstract}
This paper presents the machine learning-based ensemble conditional mean filter (ML-EnCMF) --- a filtering method based on the conditional mean filter (CMF) previously introduced in the literature.
The updated mean of the CMF matches that of the posterior, obtained by applying Bayes' rule on the filter's forecast distribution.
Moreover, we show that the CMF's updated covariance coincides with the expected conditional covariance.
Implementing the EnCMF requires computing the conditional mean (CM).
A likelihood-based estimator is prone
to significant errors for small ensemble sizes, causing the filter divergence.
We develop a systematical methodology for integrating machine learning
into the EnCMF based on the CM's orthogonal projection property.
First, we use a combination of an artificial neural network (ANN) and a linear function,
obtained based on the ensemble Kalman filter (EnKF), to approximate the CM,
enabling the ML-EnCMF to inherit EnKF's advantages.
Secondly, we apply a suitable variance reduction technique to reduce statistical errors when estimating loss function.
Lastly,  we propose a model selection procedure for element-wisely selecting the applied filter, \emph{i.e.}, either the EnKF or ML-EnCMF,
at each updating step.
We demonstrate the ML-EnCMF performance using the Lorenz-63 and Lorenz-96 systems and show that the ML-EnCMF outperforms the EnKF and the likelihood-based EnCMF.
\end{abstract}

\begin{keywords}
%\keywords{
	 nonlinear filter, inverse problem,  conditional expectation, weather forecast, deep learning
%}
\end{keywords}

\vfill
\newpage
%\tableofcontents
\input{machine_learning_da_v3}

\bibliographystyle{abbrv}
\bibliography{bibliography.bib}
\vfill % to fill the vertical space of the last page
\end{document}

%% file: machine_learning_da_v3.tex
%\usepackage{amsmath} For arxiv version
%\usepackage{graphicx}
%% ------------------------------------------------------------------
%% Code used in examples, needed to reproduce 
%% ------------------------------------------------------------------
% Macros
\newcommand\figuresfiles{figures}
\input{macros_v2.tex}

%%%%%%%%%%%%%%%%%%%%%%%%%%%%%%%%%%%%%%%%%%%%%%%%%%
\section{Introduction}\label{sec:introduction}
Data assimilation combines numerical models and observations of a dynamical system to infer its states
~\cite{reich2019data,asch2016data,reich15, Evensen2009, jazwinski1970stochastic, van2019particle}.
This approach of integrating data into dynamical models is essential in different applications,
\emph{e.g.}, numerical weather prediction, environmental forecasting, and digital twins
~\cite{asch2016data, law2015data,reichle2008data, verlaan2001nonlinearity, houtekamer1998data,bauer2021digital,
rasheed2020digital, pradhan2020global}.
Filtering is a sequential data assimilation technique, usually comprising two steps:
a) forecasting the states using the numerical models up to an observation event (prediction step) and
b) updating the forecasts by conditioning on new observations (update step).
Ideally, the update step should be performed using Bayes' rule, where the forecasting states are encoded as a prior
distribution, and the updated states are obtained as the Bayesian posterior.
However, it is computationally difficult to accurately represent the posterior distribution in the case of
nonlinear systems, even prohibitive for high-dimensional state
spaces~\cite{ades2013exploration,lei2011moment, Spantini2019}.

The ensemble Kalman filter (EnKF) extends the KF for nonlinear settings
~\cite{houtekamer1998data, Evensen2009, HLaw2016,tang2020improving, HakonHoel}.
The EnKF uses an ensemble to approximate the forecast distribution, which is then updated by applying the KF formulation
to each ensemble member.
Because the EnKF uses a linear updating formulation, the updated ensemble mean is generally biased with respect to
the Bayesian posterior mean.
Due to its linear update map, EnKF exhibits poor performance in tracking strongly nonlinear dynamical systems with sparse observations
\cite{lei2011moment, reich13, Spantini2019}.
A promising technique for developing nonlinear filters is to use methods of optimal transportation
and coupling of random variables~\cite{reich11, reich13, Spantini2019}, which is not in the scope of the paper.

The ensemble conditional mean filter (EnCMF), introduced by Lei and Bickel as
\textit{nonlinear ensemble adjustment filter} in~\cite{lei2011moment},
is a natural extension of the EnKF.
The EnCMF harnesses the advantage of the conditional expectation in characterizing the underlying conditional
distribution and ultimately improves the approximation performance of the updated ensemble for the conditional
distribution.
Compared with the \emph{ideal} Bayesian filter, the updated random variable ({\rv})
	of the conditional mean filter (CMF) does not follow the posterior
distribution obtained by applying Bayes' rule on the filter's forecast distribution~\cite{ernst2014bayesian}.
However, unlike the EnKF, the CMF's updated mean is identical to the posterior one~\cite{Matthies2016a, Vondrejc2018}.
Moreover, we show here that, though the covariance of the CMF updated RV does not match the posterior covariance,
it coincides with the expected conditional covariance.

For the practical implementation of the EnCMF,
	knowledge regarding the conditional mean (CM)
	or an accurate numerical approximation is required.
Computing the CM in nonlinear data assimilation problems with high-dimensional state vectors/fields is significantly challenging.
In~\cite{lei2011moment}, the CM is estimated using a likelihood-based approach
	via Bayes' rule,
	which is prone to significant statistical errors
	for small ensemble sizes due to the intractability of the posterior.
Here, we aim to develop a \emph{likelihood-free} nonlinear approximation for the CM
	using orthogonal projection.
Although the idea of using orthogonal projection to approximate CM
	was briefly introduced in~\cite{lei2011moment},
	the orthogonal projection-based EnCMF is implemented using only the quadratic regression
	and shows poor performance
	compared with the likelihood-based EnCMF (LL-EnCMF) in tracking Lorenz 96 system.
However, we show herein that
	by using a systematical methodology for approximating the CM,
	including variance reduction and model selection procedures,
	the proposed machine learning-based ensemble conditional mean filter
	(ML-EnCMF) outperforms the LL-EnCMF particularly for small ensemble sizes as well as the EnKF.

We use artificial neural networks (ANNs) to approximate the conditional expectation in this work.
Notably, we approximate the CM by combining the linear approximation
   obtained based on the EnKF with an ANN to represent the CM nonlinearity.
With this combination, the proposed ML-EnCMF naturally inherits
	the advantage of the EnKF for data
	assimilation problems with \emph{closely}
	linear dynamical models using small ensemble sizes~\cite{Evensen2009}.

The forecast ensemble is used as the dataset for training the ANN.
Approximating the CM using small-sized ensembles can lead to overfitting,
	which causes divergence of the filter.
We develop suitable variance reduction and
   model selection procedures to address this issue.
The applied variance reduction method aims to reduce
	statistical errors when estimating the loss function.
The model selection procedure compares
	the EnKF and ML-EnCMF predictions using
	an \emph{apriori mean squared error} (MSE)
	metric to select element-wisely the better one at each updating step.
Notably, the performance of the ML-EnCMF is generally better than the EnKF,
owing to the model selection procedure.
The effectiveness of the ML-EnCMF is demonstrated through
	detailed numerical studies of tracking the Lorenz-63 and Lorenz-96 systems
	in their chaotic regime.
Especially for the Lorenz-96 system, whose state vector is 40 dimensional,
we propose a \emph{localized} ANN structure to avoid overestimated long-range correlation
and reduce computational costs for ANN training.

The remainder of this paper is organized as follows.
In Sec.~\ref{sec:data_assimilation}, we summarize the KF, its ensemble version, and the Bayesian filter.
In Sec.~\ref{sec:CEMF}, the theoretical property of the EnCMF is discussed.
In Sec.~\ref{sec:DLEnCEMF}, the ML-EnCMF is presented.
In Sec.~\ref{sec:numerical_examples}, the performance of the ML-EnCMF implemented for tracking the Lorenz systems is
analyzed.
Finally, in Sec.~\ref{sec:conlusion}, conclusions and future research directions are discussed. %\sidenote{14.02 done}\

%%%%%%%%%%%%%%%%%%%%%%%%%%%%%%%%%%%%%%%%%%%%%%%%%%
\section{Filtering method}\label{sec:data_assimilation}
%{Data assimilation techniques seek an optimal forecast of the dynamical system states by combining modeling predictions
%with observations under the presence of uncertainty.}
This section presents the framework of the filtering approach for data assimilation.
The detailed theoretical results on filtering techniques can be found
in~\cite{reich2019data, reich15, asch2016data,Evensen2009,vetra2018state,law2015data}.

Considering the discrete-time filtering problem, which presents
the evolution of a physical system from time $t_{k-1}$ to $t_k$
by a dynamical state equation:
\begin{equation}\label{eq:dynamical_models}
	\rvQ_k = \dmodel_k(\rvQ_{k-1})\;,\quad k\in \{1,2, \dots, \kappa\}, \; \kappa \in \sN, \\
\end{equation}
where $\rvQ_k$ is an $\sR^n$-valued {\rv } of the model's states at time $t_k$, and
$\dmodel_k$ is the corresponding dynamical operator.
In practice, operator $\dmodel_k$ can be the (numerical) solution operator of a deterministic dynamical system
or a stochastic ordinary differential equation system~\cite{pathiraja2020mckean}.
The distribution of the {\rv} $\rvQ_k$ is interpreted as an uncertainty model of the actual states, denoted
as $\vqtrue(t_k)$.
Observations are recorded at time steps $t_k$ ($k\in \{ 1,2, \dots, \kappa\}$)
and modeled as
\begin{equation}\label{eq:observation}
	\rvY_k = h(\rvQ_k) + \rvXi_k,
\end{equation}
where $\rvY_k$ and $\rvXi_{k}$ are the $\sR^m$-valued {\rv s} of the observations and measurement errors, respectively,
and $h: \sR^n \rightarrow \sR^m $ is a known observation map. In practice, the observations can be spatio-temporally sparse.
We assume that the measurement error {\rv s} $\rvXi_{k}$, with $k\in \{ 0,1, \dots, \kappa\}$,
are statistically independent, the sequence of {\rv s}  $\rvQ_0,\rvQ_1, \dots, \rvQ_{\kappa}$ satisfies the Markov property,
and {\rv s} $\rvXi_{k}$ and $\rvQ_k$ are also assumed to be statistically independent.
Additionally, the distributions of the {\rv s} $\rvQ_k$, $\rvXi_{k}$, and $\rvY_{k}$ are assumed absolutely continuous,
\emph{i.e.}, their densities exist.

Let $\yob_k\in \sR^m$ denote the observation data observed at time $t_k$.
We denote the set of observation data up to time $t_k$ as $\sYob_k = \{\yob_1,\yob_2, \dots, \yob_k\}$.
An accurate representation of the  conditional probability density function (PDF) $\pi_{\rvQ_k \vert {\sYob_{k}}}$
is not feasible for high-dimensional state dynamical systems.
Based on the assumptions that $\rvQ_0,\rvQ_1, \dots, \rvQ_{\kappa}$ is a Markov process
and the measurement error {\rv s} are statistically independent,
the conditional PDF $\pi_{\rvQ_k \vert {\sYob_{k}}}$
can be sequentially approximated~\cite{Evensen2009,reich15}.

The \emph{ideal} Bayesian filtering procedure from time  $t_{k-1}$ to $t_k$ comprises two steps:
a \emph{prediction} step, where the conditional PDF $\pi_{\rvQ_{k-1} | \sYob_{k-1}}$
is transformed to the forecast PDF  $\pi_{\rvQ_{k} | \sYob_{k-1}}$ using the operator $\dmodel_k$ in Eq.~(\ref{eq:dynamical_models}),
and an \emph{update} step,
where Bayes' rule is used to map $\pi_{\rvQ_{k} | \sYob_{k-1}}$ to $\pi_{\rvQ_{k} | \sYob_k}$ as
\begin{equation}\label{eq:bayesian_posterior_PDF_global}
	\pi_{\rvQ_{k} | \sYob_{k}}(\vq)
= \dfrac{\pi_{\rvQ_{k}| \sYob_{k-1}}(\vq) \; \pi_{\rvXi_k} (\yob_k - h(\vq))}
{\int_{\sR^n} {\pi_{\rvQ_{k}| \sYob_{k-1}}(\vq') \; \pi_{\rvXi_k}(\yob_k - h(\vq'))} \;\dint \vq'},
\end{equation}
where $\pi_{\rvXi_k}$ is the PDF of {\rv } $\rvXi_k$.

Ensemble-based filters represent the uncertainty of the states via ensembles.
At the prediction step, the \emph{forecast ensembles} of the states and observations at time $t_k$,
denoted as $\{\vq_{k}^{f(i)}\}_\itn$ and $\{\vy_{k}^{f(i)}\}_\itn$,
are computed respectively as
\begin{equation}\label{eq:forecast}
\begin{aligned}
\vq_{k}^{f(i)} &= {\dmodel}_k (\vq_{k-1}^{a(i)}), \quad i= 1,2,\dots, N, \\
\vy_{k}^{f(i)} &= h(\vq_{k}^{f(i)}) + \vxi_{k}^{(i)}, \quad i= 1,2,\dots, N, \\
\end{aligned}
\end{equation}
where $N$ is the ensemble size, $\{\vq_{k-1}^{a(i)}\}_\itn$ is the \emph{{\analysis} ensemble} obtained at time $t_{k-1}$,
and $\vxi_{k}^{(i)}$ denotes the independent and identically distributed (i.i.d.) samples of $\rvXi_k$.
Here, common meteorological notations is used~\cite{reich13}, \emph{i.e.}, the superscript f (forecast) to denote
the prior statistics, and the superscript a (analysis) to denote the posterior statistics.
Generally, the updated ensemble $\{\vq_{k}^{a(i)}\}_\itn$ are computed using a suitable map $\mathcal{T}_k$ as
\begin{equation}\label{eq:analysis_ensemble}
\vq_{k}^{a(i)} = \mathcal{T}_k (\vq_{k}^{f(i)},\; \vy_{k}^{f(i)},\; \yob_{k} )\; , \quad i =1, 2, \dots, N\;.
\end{equation}
For example, the EnKF's updating map $\mathcal{T}$ is linear.
A summary of the EnKF is given in Appendix~\ref{appendix:enkf}.

The members of the ensembles $\{\vq_{k}^{f(i)}\}_\itn$, $\{\vy_{k}^{f(i)}\}_\itn$,
and $\{\vq_{k}^{a(i)}\}_\itn$
can be seen as samples of
	{\rv s} $\rvQ^f_k$, $\rvY^f_k$, and $\rvQ^a_k$, respectively, given as
\begin{subequations}
\begin{align}
	&\rvQ_{k}^{f} = \dmodel_{k}(\rvQ_{k-1}^{a}) \;, \; k\in \{ 1,2, \dots, \kappa\}\; ,\\
	&\rvY_{k}^{f} = h(\rvQ_{k}^{f}) +\rvXi_k \;, \; k\in \{ 1,2, \dots, \kappa\}\; ,\\
	&\rvQ_k^{a} = \mathcal{T}_{k} (\rvQ_{k}^{f},\; \rvY_{k}^f,\; \yob_k)\; , \;
	k\in \{ 1,2, \dots, \kappa\} \; ,\label{eq:general_form_filter}
\end{align}
\label{eq:forecast_step}
\end{subequations}
where $\rvQ_0^{a} \equiv \rvQ_0$.

Ideally, map $\mathcal{T}_k$ is identified such that $\pi_{\rvQ_k^{a}}$
is an accurate approximation of the Bayesian posterior density $\pi_{\rvQ_k^{f}|\yob_k}$, given as
\begin{equation}\label{eq:bayesian_posterior_PDF}
	\pi_{\rvQ_{k}^f | \yob_{k}}(\vq)
= \dfrac{\pi_{\rvQ_{k}^f}(\vq) \; \pi_{\rvXi_k} (\yob_k - h(\vq))} {\int_{\sR^n} \;
\pi_{\rvQ_{k}^f}(\vq') \; \pi_{\rvXi_k} (\yob_k - h(\vq'))
	\dint \vq' },
\end{equation}
which obtained from Eq.~(\ref{eq:bayesian_posterior_PDF_global})
	by substituting PDF $\pi_{\rvQ_k|\sYob_{k-1}}$ with its approximation, \emph{i.e.},
	the forecast PDF $\pi_{\rvQ_{k}^f}$.
Indeed, when $\pi_{\rvQ_k^{a}}$ is identical to $\pi_{\rvQ_k^{f}|\yob_k}$ for every $k \in \{1, 2, \dots, \kappa\}$,
we have $\pi_{\rvQ_k^{f}} \equiv \pi_{\rvQ_{k} | \sYob_{k-1}}$ and $\pi_{\rvQ_k^{a}} \equiv \pi_{\rvQ_{k} | \sYob_{k}}$ $\forall k \in \{1, 2, \dots, \kappa\}$.

Because the EnKF's updating map $\mathcal{T}_k$ is linear, for the nonlinear setting,
\emph{i.e.}, maps $\dmodel_k$ are nonlinear,
there are biased errors between
the PDFs $\pi_{\rvQ_k^{a}}$ and $\pi_{\rvQ_k^{f}|\yob_k}$.
	Consequently, the mean and higher statistical moments of PDF $\pi_{\rvQ_k^{a}}$
do not match those of PDF $\pi_{\rvQ_k^{f}|\yob_k}$~\cite{HLaw2016}.

%%%%%%%%%%%%%%%%%%%%%%%%%%%%%%%%%%%%%%%%%%%%%%%%%%
\section{Ensemble conditional mean filter}\label{sec:CEMF}%\sidenote{08.02. done}
In this section, we analyze the conditional mean filter (CMF) and its ensemble version (EnCMF).
The orthogonal projection property of conditional expectation, which is the base of the ML-EnCMF, will be discussed in detail.
First, in
Sec.~\ref{sec:conditional_mean_variance}, we recall some essential properties of the conditional expectation tailored
for the update step of the CMF and discuss two particular cases, namely, the CM and the conditional variance.
In Sec.~\ref{sec:cmf}, the CMF and its properties are analyzed.
In Sec.~\ref{sec:en_cmf_sub}, the ensemble version of CMF is presented.
Finally, in Sec.~\ref{sec:1d}, we exemplify the EnCMF using a simple static inverse problem that highlights its
differences from the EnKF.

\subsection{Conditional mean and variance}\label{sec:conditional_mean_variance}
Because our focus is on the \emph{update} step of the filtering setting,
we ignore subscript $k$ from here onward whenever possible to simplify the notations.
We assume that the variances of the {\rv s} $Q^f$ and $Y^f$ are finite.
Let $\sigma_{\rvY^f}$ be the  $\sigma-$algebra generated by the observation {\rv} $\rvY^f$ and $r: \mathbb{R}^n \rightarrow \mathbb{R}$ be
an arbitrary function such that ${r\circ\rvQ{}^f}$ --- a {\rv} composed of the forecast state {\rv} $\rvQ^f$ and
the function $r$ --- has a finite variance.
The conditional expectation $\expectation{r \circ \rvQ^f | \rvY^f}$ is a $\sigma_{\rvY^f}$-measurable RV defined as follows:
\begin{equation}\label{eq:ce_definition}
\int_{\mathcal{B}} \expectation{r\circ\rvQ^f | \rvY^f}(\omega) \probability (\dint \omega) =
\int_\mathcal{B} r\circ\rvQ^f(\omega) \probability (\dint \omega), \quad \forall \mathcal{B} \in \sigma_{\rvY^f},
\end{equation}
where $\Omega$ and $\mathbb{P}$ are an underlying probability space and measure, respectively.
The general theoretical properties of the conditional expectation can be found in, \emph{e.g.},~\cite{Bobrowski2005a}
and~\cite[Chapter~4]{durrett2019probability}.
According to the Doob-Dynkin lemma, the conditional expectation $\expectation{r \circ \rvQ^f | \rvY^f}$, as a $\sigma_{\rvY^f}$-measurable function,
takes the form $\phi_{r\circ \rvQ^f} (\rvY^f)$ for some almost surely unique measurable function
$\phi_{r\circ \rvQ^f}$~\cite{Bobrowski2005a}.
The conditional expectation has a geometric interpretation as the $L_2$ projection of the {\rv} ${r \circ \rvQ^f}$
onto the $\sigma-$algebra generated by the observation RV $\rvY^f$.
For developing our filter, we use two particular cases of the conditional expectation: the CM as the main ingredient
of the filter and the conditional variance for analyzing the variance of the updated RVs.
\subsubsection*{Conditional mean}
The CM $\expectation{\rvQ^f| \rvY^f}$ is the vector-valued {\rv} $\expectation{\rvQ^f| \rvY^f} := \phi_{\rvQ^f}\circ \rvY^f$
for some \emph{unique} function $\phi_{\rvQ^f}:\; \sR^m\rightarrow \sR^n$.
% Using the result presented in Eq.~\eqref{eq:conditioned_expectation_equiv},
The value of map  $\phi_{ \rvQ^f}$
at $\vy$ is identical to the mean of the conditional PDF $\pi_{\rvQ^f | \rvY^f}(\cdot | {\vy})$:
\begin{equation}\label{eq:conditional_mean}
\phi_{\rvQ^f}(\vy)= \int  \vq \; \pi_{Q^f|Y^f}(\vq|\vy) \; \dint \vq.
\end{equation}
Particularly, $\phi_{\rvQ^f}(\yob)$, which is obtained by evaluating map $\phi_{\rvQ^f}$ with the observation
data $\yob$, is the mean of the posterior PDF $\pi_{Q^f|\yob}$.
Usually, the CM does not have an analytical solution and is approximated using the likelihood-based
(see Appendix~\ref{appendix:likelihood_approach})
or the orthogonal projection-based approach in the filtering context.
Our paper focuses on the orthogonal projection-based approach.

Using the orthogonal projection property, the CM can be identified as follows
\begin{equation}\label{eq:cm_orthogonal_projection}
%\expectation{\rvQ^f| \rvY^f} = \phi_{ \rvQ^f}  (\rvY^f), \quad \text{where} \quad
\phi_{\rvQ^f}  =  \arg \quad \min_{g \in \mathcal{S}(\sR^m,\sR^n)}% \spaceCE}
\quad \expectation{||\rvQ^f -g \circ \rvY^f ||^2},
\end{equation}
where $\mathcal{S}(\sR^m,\sR^n)$ is the set of all functions $g:\sR^m \rightarrow \sR^n$ such that the variance of the {\rv}
$g(\rvY^f)$ is finite, and $||\cdot ||$ denotes the usual Euclidean norm. %In general, evaluating the conditional mean via the orthogonal projection requires an iterative minimizing algorithm.
When limiting the function $g$ in Eq.~\eqref{eq:cm_orthogonal_projection} to be linear, the suboptimal approximation of
the CM has a closed-form, which is given in the following lemma.

\begin{lemma}[Linear approximation of the CM] \label{lemma:linear}
The linear approximation $\linpart$ of the map  $\phi_{ \rvQ^f}$ is defined as the orthogonal projection of {\rv} $\rvQ^f$ onto
the sub-$\sigma$-algebra $\sigma_{\rvY^f}^*= \{\linpart^* \circ \rvY^f\}$ for all linear functions
$\linpart^*: \mathbb{R}^m \rightarrow \mathbb{R}^n$.
The map $\linpart$ can be analytically obtained as
\begin{equation}\label{eq:linear_approximation_ce}
\linpart (\vy) = \mat{K}\vy + \vek{b},
\end{equation}
where $\mat{K}$ is the generalized Kalman gain,
\begin{equation}\label{eq:kalman_general_form}
\mat{K}= \cov{\rvQ^f,\rvY^f}\cov{\rvY^f}^{-1},
\end{equation}
and $ \vek{b} = \expectation{\rvQ^f - \mat{K}\rvY^f}$. 
\end{lemma}
A proof of Lemma~\ref{lemma:linear} can be found in~\cite{Matthies2016a}.
Notice that, the generalized Kalman gain $\mat{K}$ defined in Eq.~(\ref{eq:kalman_general_form})
is identical to its basic version $\mat{K}^l$ (Eq.~(\ref{eq:kalman_simple_form}))
for linear observation maps, see Eq.~(\ref{eq:observation}).

\subsubsection*{Conditional variance}
The conditional covariance matrix $\cov{\rvQ^f | \rvY^f}$ is an $n\times n$-matrix valued {\rv} defined as follows
\begin{equation}
\cov{\rvQ^f | \rvY^f} \equiv \expectation{\rvQ^{f}\;\rvQ^{f \top}\; | \; \rvY^f}
-  \expectation{\rvQ^{f} \; | \; \rvY^f}
\; \expectation{\rvQ^{f} \; | \; \rvY^f}^{\top} .
\end{equation}

\subsection{Conditional mean filter}\label{sec:cmf}
The mean of the posterior PDF $\pi_{\rvQ^f | \yob}$ coincides with the CM evaluated at $\yob$, $\phi_{\rvQ^f}  (\yob)$;
hence, it is desirable to design a filter such that its updated {\rv } also has the mean value $\phi_{\rvQ^f}  (\yob)$.
This filter should also agree with the KF in the case of the linear-Gaussian setting.
The CMF proposed by Lei and Bickel in~\cite{lei2011moment} fulfills these requirements.
The CMF formulates the updated {\rv} as follows
\begin{equation}
\rvQ^a = \rvQ^f  + \phi_{\rvQ^f}  (\yob) - \phi_{\rvQ^f}  (\rvY^f).
\label{eq:ce_filter}
\end{equation}
The transformation in Eq.~\eqref{eq:ce_filter} is a nonlinear version of the general form $\mathcal{T}$ in
Eq.~(\ref{eq:general_form_filter}).
Moreover, we show here that, though the covariance of the CMF updated RV does not match the posterior covariance,
it coincides with the expected conditional covariance.
These properties of the CMF are formulated in the following theorem.

\begin{theorem}\label{theo:cmf} The updated {\rv} obtained using the CMF in
Eq.~\eqref{eq:ce_filter} satisfies the following properties:

%\begin{itemize}
A)For the linear-Gaussian setting, \emph{i.e.}, $\rvQ^f$ and $\rvXi$ are Gaussian RVs, and maps $\varPsi$and $h$ are linear,
the CMF coincides with the KF, such that
	\begin{equation}
		 \rvQ^f  + \phi_{\rvQ^f}  (\yob)  - \phi_{\rvQ^f}  (\rvY^f) \equiv \rvQ^f + \mat{K}^l(\yob - \rvY^f).
	\end{equation}%The CMF is a natural generalization of the KF filter.  Indeed, for the linear-Gaussian setting $\phi_{\rvQ^f| \rvY^f} (\rvY^f)\equiv \mat{K}\rvY^f + \vek{b}$, and the CMF  is simplified to be the KF.

B) The mean of the {\rv} $\rvQ^a$ expressed in Eq.~\eqref{eq:ce_filter} is identical to
	that of the posterior PDF $\pi_{\rvQ^f_k | \yob_k}$  (Eq.~\eqref{eq:bayesian_posterior_PDF}):
	\begin{equation}
	\expectation{\rvQ^a} =  \phi_{\rvQ^f}  (\yob),
	\end{equation}

C) The covariance of the RV $\rvQ^a$ expressed in Eq.~\eqref{eq:ce_filter}
	is equal to the expected conditional covariance:
	\begin{equation}
	\cov{\rvQ^a} = \expectation{\cov{\rvQ^f | \rvY^f}}.
	\end{equation}
%\end{itemize}
\end{theorem}

\begin{proof}
	A) For the linear-Gaussian setting, we achieve that $\phi_{\rvQ^f} (\rvY^f)\equiv \mat{K}^l\rvY^f + \vek{b}$ using
	Lemma~\ref{lemma:linear}.
	The updated {\rv} $\rvQ^a$ obtained using Eq.~(\ref{eq:ce_filter}) becomes
	$\rvQ^a = \rvQ^f + \mat{K}^l(\yob - \rvY^f)$,
	which is identical to the KF (Eq.~(\ref{eq:KF_filter_interms_rv})).\\
	B) Let $\rvQ^I$ be defined as
	$\rvQ^I := \rvQ^f - \expectation{\rvQ^f| \rvY^f}$.
	Based on the law of total expectation, we obtain the mean and variance of $\rvQ^I$ as follows:
	\begin{equation}
	\expectation{\rvQ^I} = \vzeros_n,
	\label{eq:zero_difference_of_means}
	\end{equation}
	where $\vzeros_n$ is the $n$-dimensional zeros vector, and
	\begin{equation}
	\cov{\rvQ^I} = \cov{Q^f} - \cov{\expectation{\rvQ^f| \rvY^f}},
	\end{equation}
	respectively. Thus, the mean of the updated {\rv} $\rvQ^a$ is identical to $\phi_{\rvQ^f}  (\yob)$:
	\begin{equation}
	\expectation{\rvQ^a}= \expectation{\rvQ^I} +  \phi_{\rvQ^f}  (\yob) =  \phi_{\rvQ^f}  (\yob),
	\end{equation}
	which is the mean of the posterior PDF as expressed in Eq.~(\ref{eq:conditional_mean}).\\
	C)  
	Using the law of total variance
	\begin{equation}
	\cov{Q^f} = \expectation{\cov{Q^f|\rvY^f}} +\cov{\expectation{\rvQ^f |\rvY^f}},
	\end{equation}
	we obtain
	\begin{equation}
	\cov{Q^a} = \cov{Q^f} - \cov{\expectation{\rvQ^f| \rvY^f}} =  \expectation{\cov{\rvQ^f | \rvY^f}}.
	\end{equation}
\end{proof}

Property A of Theorem~\ref{theo:cmf} indicates that the CMF is a natural extension of the KF.
Compared with the posterior PDF, $\pi_{Q^f|\yob}$,
the updated {\rv} $\rvQ^a$ of the CMF expressed in Eq.~(\ref{eq:ce_filter}) shows an identical mean vector
(Property B).
	 However, the distribution of the centered updated RV,
	 $\rvQ^f- \phi_{\rvQ^f}(\rvY^f)$, and
     the covariance matrix $\cov{\rvQ^a}$, in particular, are independent of the observational data $\yob$
	and not identical to those of the posterior PDF,
    besides special cases such as
    the linear-Gaussian setting.
     An approach to improve the CMF is to match the covariance
     and possibly higher moments of the updated RV
      with the empirical posterior moments,
    as proposed in~\cite{lei2011moment,Matthies2016a},
     which requires significant additional computational resources.
	In an extreme scenario where the conditional mean does not encode any information,
    \emph{i.e.}, the conditional mean has zero variance,
     the CMF exhibits no information gain, \emph{i.e.}, $\rvQ^a \equiv \rvQ^f$.
    This limitation also applies to the EnKF.

\begin{remark}\label{remark:covariance_matrix}
Usually, the conditional variance and its range are significantly diminished compared
with the variance of the forecast {\rv} $\rvQ^f$ in practice
because uncertainty is reduced thanks to observational data.
Hence, the error between the expected conditional covariance
and the posterior covariance $(\cov{\rvQ^f | \rvY^f=\yob})$
is much smaller than the forecast one $(\cov{\rvQ^f})$.
In other words, compared with the posterior PDF $\pi_{Q^f|\yob}$, the updated
	{\rv} $\rvQ^a$ has an \emph{approximate} covariance matrix (Theorem~\ref{theo:cmf}, Property C).
We illustrate this argument in Sec.~\ref{sec:1d}.
Moreover, Property C is advantageous when combining EnCMF with the method
of A-optimal design of experiments~\cite{AlPeStGh16Aoptiamal, crestel2017optimal},
	which seeks for the experiment setup to minimize the trace of the expected conditional covariance.
\end{remark}

\subsection{Ensemble approximation of conditional mean filter}\label{sec:en_cmf_sub}
Similar to the EnKF, the EnCMF uses  the ensemble technique to represent the forecast and updated {\rv s}.
In the forecast step, the distributions of the {\rv s} $\rvQ^f$ and $\rvY^f$ are approximated using
the $N$-sized ensembles,
$\{\vq^{f(i)}\}_\itn$ and $\{\vy^{f(i)}\}_\itn$, respectively, as expressed in Eq.~(\ref{eq:forecast_step}).
In the update step, the {\analysis} ensemble $\{\vq^{a(i)}\}_\itn$ of the {\rv} $\rvQ^a$ is
evaluated using Eq.~(\ref{eq:ce_filter}):
\begin{equation}\label{eq:en_cmf}
	\vq^{a(i)} = \vq^{f(i)} + \phi_{\rvQ^f}  (\yob) - \phi_{\rvQ^f}  (\vq^{f(i)}) , \quad i =1,\; 2, \dots,  N.
\end{equation}

Implementing the EnCMF requires the evaluation of map $\phi_{\rvQ^f}$ of the CM.
This task can be performed using two approaches: a) by evaluating the conditional PDF in Eq.~(\ref{eq:conditional_mean})
via the likelihood function
or b) using the orthogonal projection property expressed in Eq.~(\ref{eq:cm_orthogonal_projection}).
The likelihood-based approach, which uses the sampling technique to estimate values of the conditional mean for
 different samples of measurement RV $\rvY^f$
 (summarized in Appendix~\ref{appendix:likelihood_approach}),
has limited applicability for data assimilation problems with high-dimensional state-spaces
because the conditional distribution becomes intractable.
The orthogonal projection-based approach is favorable in such a situation.
In Sec.~\ref{sec:1d}, we implement the EnCMF with a simple one-dimensional example, in which the likelihood-based approach is applied.
In Sec.~\ref{sec:DLEnCEMF}, we discuss the approximation of the CM using the orthogonal projection-based approach combined with machine learning (ML),
	which will lead to the ML-EnCMF method.

Algorithm ~\ref{algorithm:meta_MLEnCMF} presents a pseudocode of the ML-EnCMF.
The forecast step is similar to those of the EnKF.
The update step using ML-based approximation of the CM will be presented in Sec.~\ref{sec:DLEnCEMF}.
In terms of execution schedule, the prediction step and the approximation of the CM do not require observational data;
hence they can be performed prior to the arrival of the data for accelerating the filtering process.

\begin{algorithm}[ht]
	\caption{ML-EnCMF algorithm for performing assimilation at time $t_k$.}
	\label{algorithm:meta_MLEnCMF}
	\begin{algorithmic}[1]
	\Require Updated ensembles obtained from the assimilation step performed at $t_{k-1}$
	$\{\vq_{k-1}^{a(i)}\}_\itn$ and observational data $\yob_k$
	\Statex
	\Statex \textbf{\textit{Prediction}}:\;
	\State Evaluate forecast ensembles $\{\vq_k^{f(i)}\}_\itn$  \Comment Eq.~(\ref{eq:forecast})
	\Statex
	\Statex \textbf{\textit{Update}}:\;
	\State Approximate the map $\phi_{\rvQ_k^f}$ (explained in Sec.~\ref{sec:DLEnCEMF})
		\Comment Algorithm~\ref{algorithm:meta_training}

	\State Evaluate the updated ensemble $\{\vq_k^{a(i)}\}_\itn$ \Comment Eq.~(\ref{eq:ml_encmf})
	\Statex
	\Statex
	\Return Updated ensemble $\{\vq_k^{a(i)}\}_\itn$
	\end{algorithmic}
\end{algorithm}

\subsection{Illustration of the EnCMF  for a simple static inverse problem}\label{sec:1d}
We consider the following inverse problem to compare the EnCMF with the EnKF and the ideal Bayesian filter.
Given a one-dimensional {\rv} $\rvQ^f\sim\mathcal{N}(0,2^2)$ and the following nonlinear observation map,
\begin{equation}\label{eq:1d_example}
\rvY^f = h(\rvQ^f) + \rvXi \quad \text{where} \quad h(q)=
\begin{cases}
q   &\quad \textnormal{if}\;  q\leq 0\\
q^2 &\quad \textnormal{if}\; q > 0
\end{cases}, \quad\rvXi \sim \mathcal{N}(0, 0.5^2),
\end{equation}
the task is to evaluate the Bayesian posterior for different values of the observation $\yob$.
We solve the problem using the EnCMF and {EnKF} and compare their updated ensembles with the Bayesian posterior in terms of three statistical characteristics: mean, variance,
and PDF.
The observation map considered here is nonlinear;
therefore, we employ Eq.~(\ref{eq:kalman_general_form}) for computing Kalman gain.
For this simple problem, we use the likelihood-based approach, explained in Appendix~\ref{appendix:likelihood_approach},
for approximating the conditional mean (Eqs.~(\ref{eq:ll_cm},~\ref{eq:ll_cm_next})) and implementing the EnCMF (Eq.~(\ref{eq:ll_encmf})).
A large ensemble size of 10000 is selected to minimize statistical errors.

\begin{figure}[th!]
	\centering 
	\includegraphics[scale=0.37]{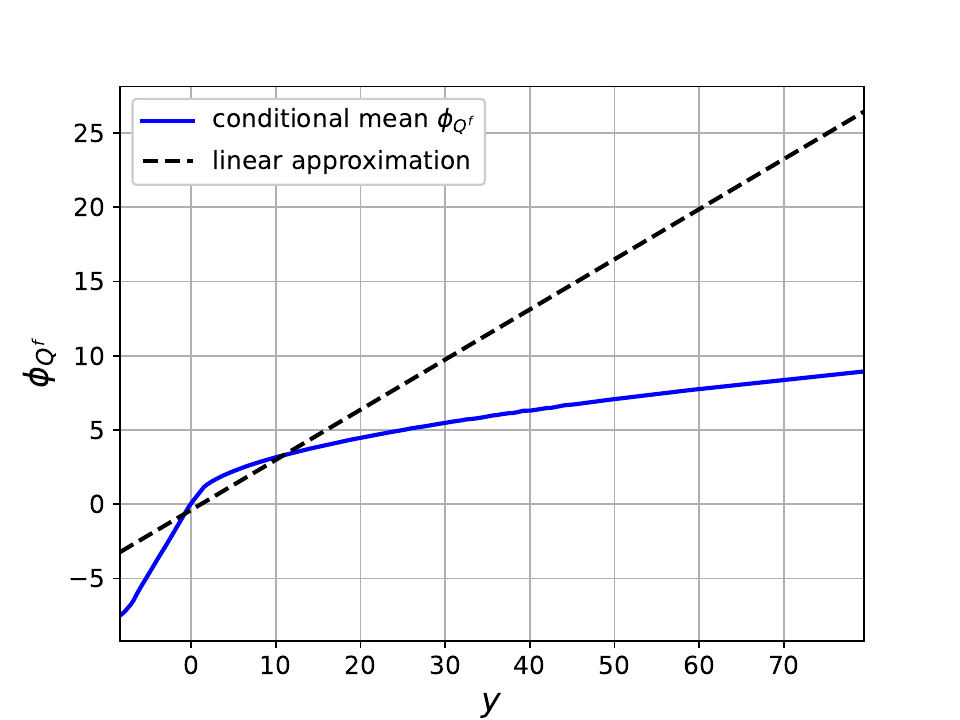}
	\caption{CM computed using Eq.~(\ref{eq:conditional_mean}) and its linear approximation (Eq.~(\ref{eq:linear_approximation_ce})).}
	\label{fig:ce_1d}
\end{figure}

We refer to Eq.~(\ref{eq:kalman_general_form}) and apply the MC method with the same ensemble size to approximate the Kalman gain.
For comparison, Fig.~\ref{fig:ce_1d} depicts the CM map $\phi_{ \rvQ^f}$ with its linear approximation, obtained using Eq.~(\ref{eq:linear_approximation_ce}).
We observe a significant error in the linear approximation.
Consequently, a significant bias between the means of the updated ensemble of the {EnKF} and the Bayesian posterior is predicted. %\sidenote{12:02: done}

\begin{figure}[h!]
	\centering 
	\includegraphics[scale=0.37]{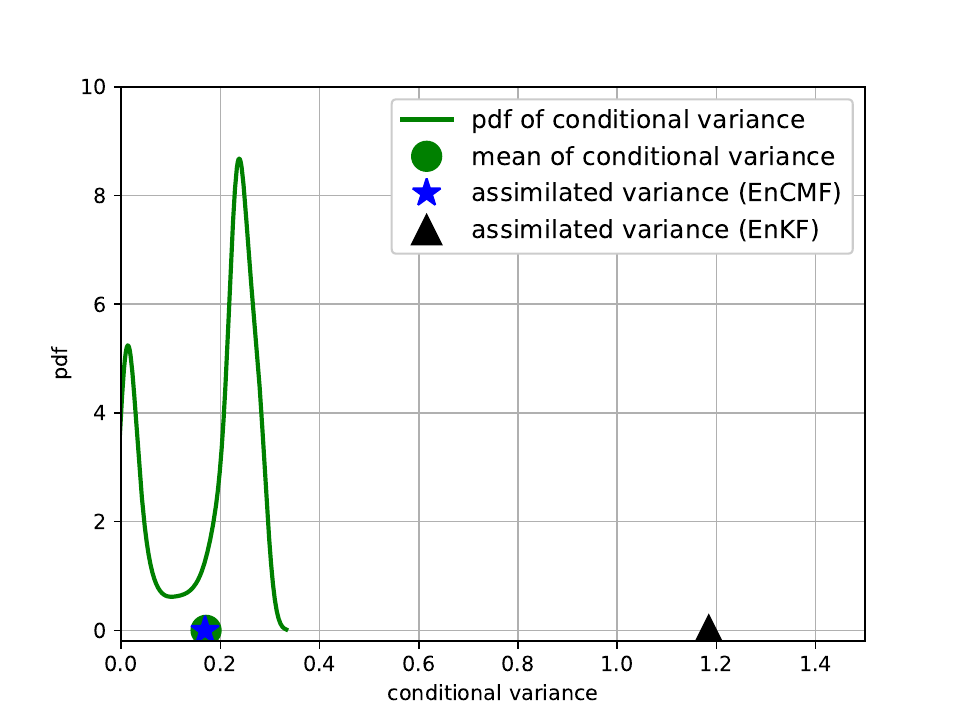}
	\caption{Empirical PDF of the conditional variance $\variance{Q^{f }|Y^f}$ expressed in  Eq.~(\ref{eq:conditional_variance_simple1d}b) compared with the variances of the updated ensembles obtained using the EnKF and EnCMF. The expected conditional variance, $\expectation{\variance{Q^{f }|Y^f}}$,  and the updated ensemble variance of the EnCMF show closely identical estimated values of approximately 0.17.  }
	\label{fig:cvariance_1d}
\end{figure}

We approximate the conditional expectation of the second moment $\expectation{\left(Q^{f} \right)^2|Y^f}$
and the conditional variance $\variance{Q^{f }|Y^f}$ using the conditional PDF as follows
\begin{subequations}\label{eq:conditional_variance_simple1d}
\begin{align}
&\expectation{\left(Q^{f} \right)^2|Y^f= y} = \int q^2 \pi_{\rvQ^f | \rvY^f}(q|y) \dint q,\\
&\begin{aligned}
\variance{Q^{f }|Y^f} &= \expectation{\left(Q^{f} \right)^2|Y^f)} - \left (\expectation{Q^f|Y^f)} \right)^2 \\
&= \expectation{\left(Q^{f} \right)^2|Y^f} - \left (\phi_{ \rvQ^f}(\rvY^f) \right )^2, 
\end{aligned}
\end{align}
\end{subequations}
where the likelihood-based approach is applied to compute $\expectation{\left(Q^{f} \right)^2|Y^f= y}$,
see Eq.~(\ref{eq:ll_cm2}). Fig.~(\ref{fig:cvariance_1d}) presents the empirical PDF of the conditional variance.

We implement the {EnKF}, using Eqs.~(\ref{eq:forecast},~\ref{eq:kalman_general_form},~\ref{eq:enkf}), and the LL-EnCMF,
using Eqs.~(\ref{eq:forecast}, \ref{eq:ll_cm_next}, \ref{eq:ll_encmf}).
The updated ensemble variances in both filters are invariant
when varying measurement data.
For the EnCMF, the updated ensemble variance is a non-biased estimator of the mean of
the conditional variance; thus, the EnCMF shows significant improvement in estimating the conditional variance
than the {EnKF}.
Notably, the absolute error between the EnCMF's updated variance and the conditional variance (within $[0, 0.18)$)
is much smaller than the prior variance (equal to 4).
This observation is an illustration for the Remark~\ref{remark:covariance_matrix}.

\begin{figure}[h!]
	\centering
	\begin{subfigure}[b]{0.45\textwidth}
	    \centering
		\includegraphics[scale=0.35]{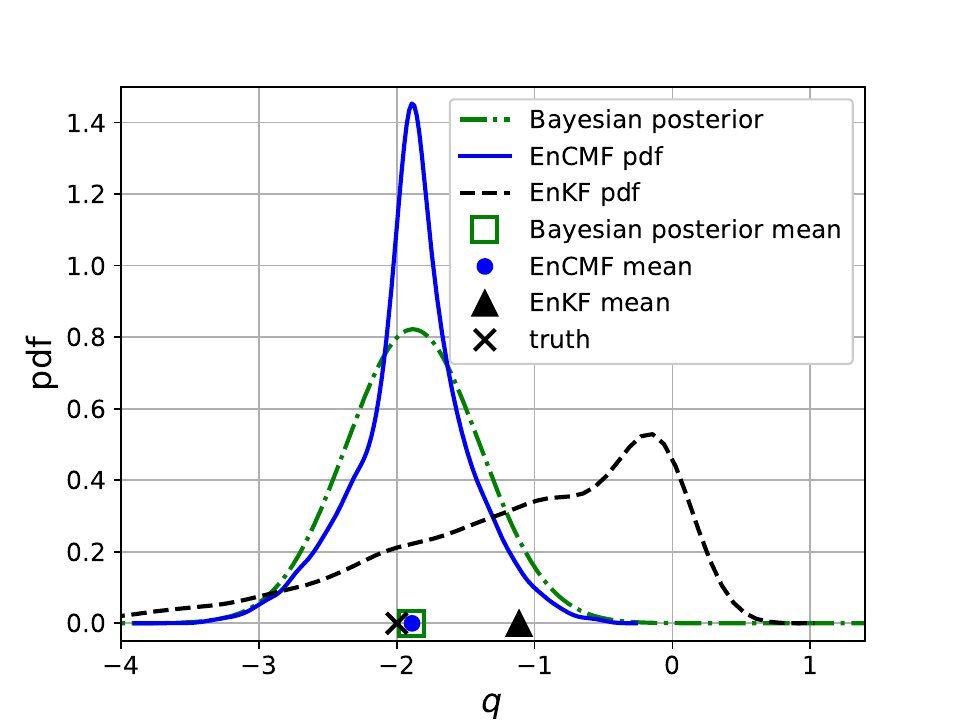}
		\caption{$~$}
	\end{subfigure}
	\begin{subfigure}[b]{0.45\textwidth}
	    \centering
		\includegraphics[scale=0.35]{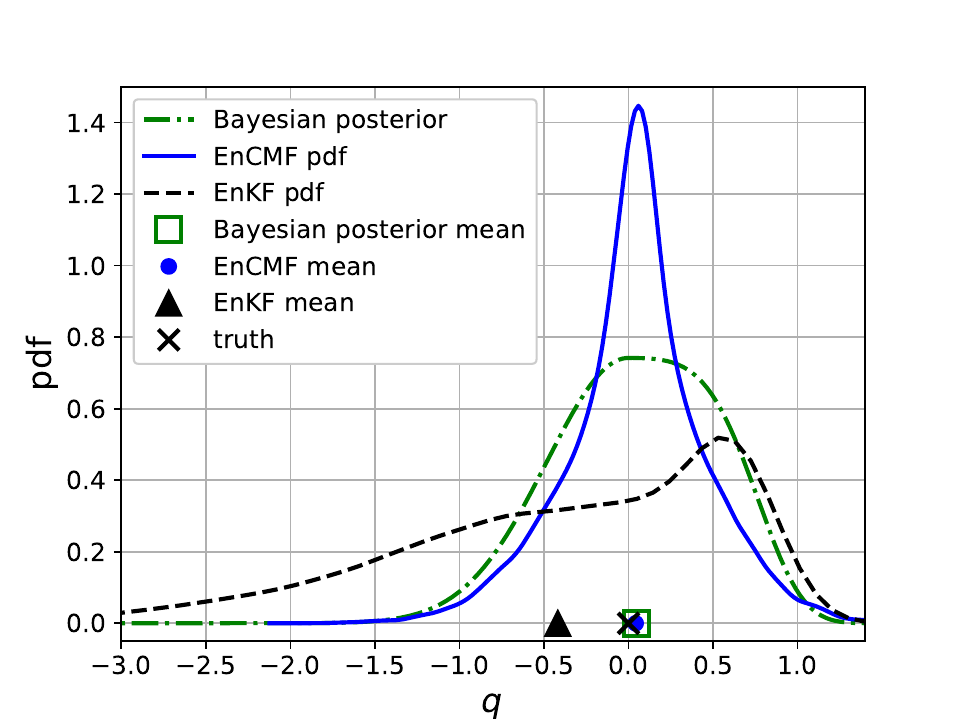}
		\caption{$~$}
	\end{subfigure}

	\begin{subfigure}[b]{0.5\textwidth}
	    \centering
		\includegraphics[scale=0.35]{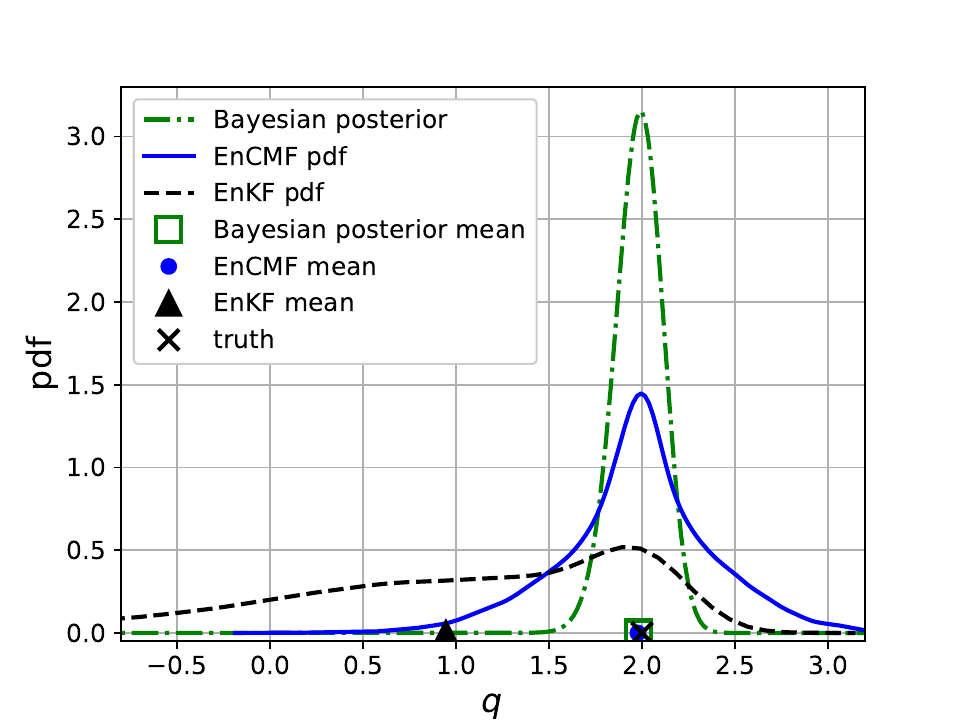}
		\caption{$~$}
	\end{subfigure}
	\caption{Comparison of the empirical densities of the updated ensembles and the Bayesian posterior:
	 (a) actual value $\qtrue = -2$ (minus standard deviation of the prior distribution), (b) actual value $\qtrue = 0$ (mean of the prior distribution), (c) actual value $\qtrue = +2$ (plus standard deviation of the prior distribution).}
	\label{fig:enkf_encmf_truth}
\end{figure}

We consider different observation scenarios using various $\qtrue$ values to evaluate the \textit{synthesized}
observation data as $\yob = h(\qtrue) + \xi$, where $\xi$ is an i.i.d. sample of the error {\rv} $\rvXi$.
Fig.~\ref{fig:enkf_encmf_truth} presents the empirical densities of updated ensembles and the posterior PDF.
Although the empirical PDF of the updated ensemble obtained using the EnCMF
does not coincide with the Bayesian posterior,
it still fits the posterior significantly better than the {EnKF}.
In particular, the updated ensemble mean obtained using the EnCMF is closely identical to the Bayesian posterior mean
and in good agreement with the truth value $\vqtrue$.
In contrast, the updated ensemble of the EnKF exhibits significant biased errors
in terms of the mean value compared with the Bayesian posterior.
This numerical experiment reconfirms the theoretical statement in Theorem~\ref{theo:cmf}.%\sidenote{12:02 done}

A counter-example to illustrate the limitation of  the CMF,
 discussed in Sec.~\ref{sec:cmf},
  is to replace the observation map in Eq.~(\ref{eq:1d_example})
   by $h=q^2$ \cite{ernst2014bayesian}.
In this case, the CMF fails to incorporate measurement data
because the conditional mean is deterministic
$\phi_{Q^f}(Y^f) = 0$.
Indeed the updated RV of the CMF is identical to the forecast.

		%%%%%%%%%%%%%%%%%%%%%%%%%%%%%%%%%%%%%%%%%%%%%%%%%%
\section{ML-based EnCMF}\label{sec:DLEnCEMF}
This section presents the ML-EnCMF. While our paper focuses on the update step,
improving the efficiency of the forecast step using ML techniques is also actively investigated, \emph{e.g.}, in
\cite{abarbanel2018machine,bocquet2020bayesian,brajard2020combining, gagne2020machine}.
Based on the orthogonal projection property (Eq.~(\ref{eq:cm_orthogonal_projection})),
we develop an approximation of the CM using ANN.
The updated ensemble is obtained by inserting the CM approximation in Eq.~(\ref{eq:en_cmf}).

The section is organized as follows. In Sec.~\ref{sec:ANN_approximation},
we discuss the approximation of the CM using a combination of the KF linear updating map
and an ANN.
In Sec.~\ref{sec:data_augmentation} and Sec.~\ref{sec:cross_validation}, the variance reduction
and the model selection methods are respectively reported.
Finally, Sec.~\ref{sec:updating_algorithm} reports a pseudo-algorithm for the training and model selection procedures.

\subsection{ANN-based approximation of the conditional mean}\label{sec:ANN_approximation}
	Let $\nlinpart(\cdot ; \weight): \mathbb{R}^m \rightarrow \mathbb{R}^n$ be an ANN map, where $\weight$ denotes
its hyper-parameters, \emph{i.e.}, the network's weights and biases.
The ANN structure is not fixed and adapted following the natural representation of the states.
For example, convolution neural networks can be used for spatial/field states to represent the spatial correlation,
as illustrated in Sec.~\ref{sec:numerical_results:lorenz96}.
 Here, we approximate the CM, \emph{i.e.}, map $\phi_{\rvQ^f}$, by combining the linear approximation $g_{\linear}$
(Eq.~(\ref{eq:linear_approximation_ce})) and the ANN $\nlinpart(\cdot ; \weight)$ as
\begin{equation}
 	\phi_{\rvQ^f}(\cdot) \approx \linpart (\cdot)+ a \odot \; \nlinpart(\cdot ; \weight),
 	\label{eq:cm_approximation}
 \end{equation} 
where $a=[a_1,\dots, a_n]$ is a $n$-dimensional vector with $a_{\alpha} \in \{0, 1\}$ with $\alpha = 1, \dots, n$,
	and $\odot$~denotes the element-wise multiplication.
	In other words, the $\alpha$\textit{th}-component of the state vector is updated using
	either the combined nonlinear map ($a_{\alpha} = 1$) or the Kalman linear map ($a_{\alpha} =0$).
	The value of vector $a$ is identified based on the model selection described in Sec.~\ref{sec:cross_validation} below.

Instead of relying solely on the ANN to approximate the map $\phi_{\rvQ^f}$, the combination
in Eq.~(\ref{eq:cm_approximation}) is more robust.
Indeed, the proposed combination inherits from the acceptable performance of the EnKF in closely linear-Gaussian settings with small ensemble sizes.
Performance-enhancing techniques successfully applied for the EnKF with small-sized ensembles,
\emph{e.g.}, covariance tapering~\cite{gaspari1999construction},
are also applicable for the linear part of the ML-EnCMF.
Moreover, the ML-EnCMF generally performs better than the EnKF
	because the filter is only activated when it approximates the CM better than the EnKF,
owing to the model selection procedure.

After the ANN $\nlinpart$ is trained, given the observation data $\yob$, the updated {\rv} $\rvQ^a$ of the CMF (Eq.~(\ref{eq:ce_filter})) and its ensemble (Eq.~(\ref{eq:en_cmf})) are approximated by the following relations
\begin{subequations}
\begin{align}
\rvQ^a &= \rvQ^f + \mat{K}  (\yob-\rvY^f) + a \odot \bigl (\nlinpart(\yob;\weight) -\nlinpart(\rvY^f;\weight)  \bigr), 	\label{eq:ce_filter_ML}\\
\vq^{a (i)} &= \vq^{f(i)} + \mat{K}  (\yob- \vy^{f (i)}) +a \odot \bigl (\nlinpart(\yob;\weight) - \nlinpart(\vy^{f (i)};\weight) \bigr),
\label{eq:ml_encmf}
 \end{align}
\end{subequations}
where $i =1, \dots, N$, respectively.
The mean and covariance estimators of the updated ensemble are given as
\begin{subequations}
\begin{align}
    &\phantom{\cov{\rvQ^a}} \expectation{\rvQ^a} \approx \linpart(\yob) + a \odot \nlinpart(\yob;\weight), \\
	&\begin{aligned}
	\phantom{\expectation{\rvQ^a}} \cov{\rvQ^a} \approx \dfrac{1}{N}\sum_{i=1}^N &\bigl[\vq^{f(i)} - \linpart(\vy^{f (i)})
	- a \odot \nlinpart(\vy^{f (i)} ;\weight)\bigr]\\
													&\bigl[\vq^{f(i)} - \linpart(\vy^{f (i)})
	- a \odot \nlinpart(\vy^{f (i)} ;\weight)\bigr]^\top,
	\end{aligned}\label{eq:updated_var}
\end{align}
\end{subequations}
using Theorem~\ref{theo:cmf}.

We train the map $\nlinpart$ by using the orthogonal projection property of the CM stated in Eq.~(\ref{eq:cm_orthogonal_projection}).
For a given tensor of hyper-parameters $\weight$, we use the MSE $\mse (\weight)$ defined as
\begin{equation}
\mse (\weight) = \expectation{\bigl \lVert \rvQ^f - \linpart (\rvY^f)-\nlinpart(\rvY^f; \weight)  \bigr \rVert^2}
\end{equation}
as a metric for the approximation in Eq.~(\ref{eq:cm_approximation}).
The hyper-parameters of the ANN are obtained by solving the following optimization problem
\begin{equation}
\weight = \arg \quad \min_{\weight_{*}} \qquad {\mse}(\weight_{*})+ \mathcal{R}(\weight_{*}),%| \mathcal{D}). %+  Re(\weight, \lambda) ,
\label{eq:optimization_problem}
\end{equation} 
where $\mathcal{R}(\weight_{*})$ is a regularization term applied to reduce overfitting, usually composed of $L_1$ or $L_2$ norms of the weights \cite{burden2008bayesian}. 

\begin{remark}
An important metric to assess the performance of a filter is the $L_2$ norm of the error between the updated ensemble mean and
the ground-truth state $\vq^{\textnormal{tr}}$. For example, that metric for the ML-EnCMF is evaluated as $\Vert \vq^{\textnormal{tr}} -
\linpart (\vy_{\textnormal{obs}})-\nlinpart(\vy_{\textnormal{obs}}; \weight) \Vert^2$ supposing vector $a$ is a unit one.
Therefore, the value of $\mse (\weight)$ is \emph{the apriori MSE} of the updated ensemble mean.
Furthermore, $\mse (\weight)$ is equal to the total variance of the updated ensemble (see Eq.~\eqref{eq:updated_var}),
	which is a measure of the prediction uncertainty.
In other words, minimizing the function $\mse$ is equivalent to minimizing
the apriori MSE of the updated ensemble mean
and simultaneously minimizing the prediction uncertainty.
\end{remark}

\subsection{Reduced variance estimator of the MSE metric}\label{sec:data_augmentation}
In our approach, the metric ${\mse}(\weight)$ is estimated using the forecast ensembles. 
Let $\mathcal{D}=\{\bigl(\vy^{f(i)},\; \vq^{f(i)}\bigr)\}_\itn$ be the dataset comprising
pairs $\bigl(\vy^{f(i)},\; \vq^{f(i)}\bigr)$ collected from the forecast ensembles,
	evaluated using Eq.~(\ref{eq:forecast}) in the prediction step.
The dataset $\mathcal{D}$ is divided into two sets:
an $N_T$-sized training dataset $\mathcal{D}_T$ for tuning the hyper-parameters
and an $N_S$-sized test dataset $\mathcal{D}_S$ for testing the ANN; here, $N_T + N_S = N$.
In both training and testing processes, the metric $\mse(\weight)$ is estimated based on the corresponding datasets, $\mathcal{D}_T$ and $\mathcal{D}_S$, respectively.
Statistical errors are inherent in such an estimation;
hence, we present a variance reduction technique to reduce these errors.

The crude MC estimator of the metric
$\mse$ using the training dataset is obtained as follows
\begin{equation}\label{eq:MC_estimator}
\widehat{\mse}(\weight | \mathcal{D}_T) = \dfrac{1}{N_T}\sum_{(\vy^{f(i)},\; \vq^{f(i)})\in \mathcal{D}_T}
\bigl \lVert \vq^{f(i)}- \linpart (\vy^{f(i)}) -\nlinpart(\vy^{f(i)}; \weight) \bigr \rVert^2,
\end{equation}
where $\vy^{f(i)} = h(\vq^{f(i)}) +\vxi^{(i)}$ as stated in Eq.~(\ref{eq:forecast}).
We use the following reduced-variance estimator obtained by suitably augmenting the data to reduce the statistical errors in evaluating the MSE metric

\begin{equation}\label{eq:MC_reduced_variance}
\begin{aligned}
\widehat{\mse}^{\textnormal{vr}}(\weight | \mathcal{D}_T) = \dfrac{1}{N_T } \sum_{(\vy^{f(i)},\;\vq^{f(i)})\in \mathcal{D}_T}
\dfrac{1}{M} \sum_{j=1}^{M} \bigr \rVert \vq^{f(i)}
&- \linpart \bigl(h(\vq^{f(i)})+\vxi^{(i,j)}\bigr)\\
&- \nlinpart\bigl(h(\vq^{f(i)})+\vxi^{(i,j)}; \weight\bigr) \bigr \rVert^2,\\
\end{aligned}
\end{equation}
where $M$ is the data augmentation multiplier, and $\vxi^{(i,j)}$ with $i=1,2,\dots,N$ and $j =1,2,\dots,M$
	are i.i.d. samples of {\rv} $\rvXi$.
When $M$ is sufficiently large, the estimator $\widehat{\mse}^{\textnormal{vr}}(\weight | \mathcal{D}_T)$
shows minor statistical errors compared with $\widehat{\mse}(\weight | \mathcal{D}_T)$
(see discussion in Appendix~\ref{appendix:variance_duction}). %\sidenote{24.02 done}

Intuitively, this variance reduction technique decreases the sensitivity of the trained ANN to observation noise because additional noisy data are used for training.
From the implementation perspective, using the reduced-variance estimator
$\widehat{\mse}^{\textnormal{vr}}(\weight | \mathcal{D}_T)$
is equivalent to training the ANN on an $N \times M$ augmented dataset:
 \begin{equation}\label{eq:augdata}
 \mathcal{D}^a=\{(\vy^{f(i,j)},\; \vq^{f(i)}), \quad i=1,\dots,N, \; j = 1,\dots, M \},% \quad \textnormal{where} \; \vy^{f(i,j)} = h(\vq^{f(i)}) +\vxi^{(i,j)},
 \end{equation}
where $\vy^{f(i,j)} = h(\vq^{f(i)}) +\vxi^{(i,j)}$, which does not require any modification of the implemented training algorithm.

\subsection{Model selection}\label{sec:cross_validation}%\sidenote{08.02, 4.3 done}
Training the ANN with possibly many hyper-parameters on small datasets can lead to overfitting.
%To alleviate this risk, we only use the high-dimensional and complex ANN model if it significantly improves
%for the goodness-of-fit.
Although the CM may be nonlinear, the linear approximation (Eq. (\ref{eq:linear_approximation_ce}))
may still be preferable over an ANN-based approximation that is significantly overfitted
owing to robustness.
Our proposed model selection compares the ML-EnCMF with the EnKF
using the apriori MSE of the updated ensemble mean.

Let $L_{\alpha}$ and $J_{\alpha}$, $\alpha=1,\dots, n$,
	be the apriori MSE of the EnKF and ML-EnCMF ensemble means, respectively,
of the $\alpha-th$ state component, defined as
\begin{subequations}
	\begin{align}
L_{\alpha} &= \expectation{\bigl([\;\rvQ^f]_\alpha - [\linpart (\rvY^f)]_{\alpha}\;\bigr)^2},\\
J_{\alpha} &= \expectation{\bigl([\;\rvQ^f]_\alpha - [\linpart (\rvY^f)]_{\alpha} -  [\nlinpart(\rvY^f; \weight)]_{\alpha}\;\bigr)^2}.
	\end{align}
\end{subequations}
We identify the vector $a$ in Eq.~(\ref{eq:cm_approximation}) as
\begin{equation}\label{eq:model_selection}
a_{\alpha} = \mathbf{1} (\widehat{L}_{\alpha}> \widehat{\mse}_{\alpha}), \quad \alpha=1,\dots, n
\end{equation}
where $\mathbf{1}(\cdot)$ is a logical operator yielding one if the condition $(\cdot)$ is valid and zero otherwise.
In Eq.~(\ref{eq:model_selection}), $\widehat{L}_{\alpha}$ and $\widehat{J}_{\alpha}$ are estimated, respectively, as:
\begin{subequations}\label{eq:validation_metric}
\begin{align}
&\widehat{L}_{\alpha}= \dfrac{1}{N_S}\sum_{(\vy^{f(i)},\; \vq^{f(i)})\in \mathcal{D}_S} \dfrac{1}{M}\sum_{j=1}^{M}
\biggl( \bigl[\vq^{f(i)}\bigr]_{\alpha} - \bigl[\linpart \bigl (h(\vq^{f(i)})+ \vxi^{(i,j)} \bigr)\bigr]_{\alpha} \biggr)^2, \label{eq:model_slection_metric_linear}\\
&\begin{aligned}
\widehat{J}_{\alpha}= \dfrac{1}{N_S}\sum_{(\vy^{f(i)},\; \vq^{f(i)})\in \mathcal{D}_S} \dfrac{1}{M}\sum_{j=1}^{M}
\biggl( \bigl[\vq^{f(i)}\bigr]_{\alpha}&- \bigl[\linpart \bigl (h(\vq^{f(i)}) +\vxi^{(i,j)} \bigr )\bigr]_{\alpha} \\
&- \bigl[\nlinpart\bigl (h (\vq^{f(i)}) +\vxi^{(i,j)};\weight \bigr)\bigr]_{\alpha} \biggr)^2,
\label{eq:model_slection_metric_ann}
\end{aligned}
\end{align}
\end{subequations}
where we apply the variance reduction technique, similar to Eq.~(\ref{eq:MC_reduced_variance}),
	and use the test dataset to avoid bias errors.

\subsection{Algorithm}\label{sec:updating_algorithm}
Given the forecast ensembles $\{\vq_k^{f(i)}\}_\itn$, $\{\vy_k^{f(i)}\}_\itn$, we estimate the Kalman gain,
using Eq.~(\ref{eq:kalman_simple_form}) (Eq.~(\ref{eq:kalman_general_form}))
for linear (nonlinear) observational maps, accordingly.
We then approximate the map $\phi_{\rvQ_k^f}$ as stated in Eq.~(\ref{eq:cm_approximation}).
The training pseudocode is presented in Algorithm~\ref{algorithm:meta_training}.
Using the forecast ensemble, $\mathcal{D}=\{\bigl(\vy_k^{f(i)},\; \vq_k^{f(i)}\bigr)\}_\itn$,
we first generate the augmented dataset (Eq.~(\ref{eq:augdata})),
which is used to compute the reduced-variance estimators of the MSE metric as expressed
in Eqs.~(\ref{eq:MC_reduced_variance},~\ref{eq:validation_metric}).
Then, to solve the MSE minimizing problem described in Eq.~(\ref{eq:optimization_problem}),
this algorithm uses the mini-batch version of the stochastic gradient descent method,
a widely accepted approach for ANN training~\cite{kingma2014adam}.
We include a call-back procedure into the training process to select the hyper-parameters with the best performance~\cite{Goodfellow2016a}.
Finally, the model selection procedure (explained in Sec.~\ref{sec:cross_validation}) is performed.
Algorithms ~\ref{algorithm:meta_MLEnCMF} and~\ref{algorithm:meta_training},
	together, summarize our ML-EnCMF implementation.

\begin{algorithm}[H]
	\caption{Training algorithm of the ANNs used for approximating the map $\phi_{ \rvQ^f}$.}
	\label{algorithm:meta_training}
	\begin{algorithmic}[1]
	\Statex
	\Require {Ensembles $\{\vq^{f(i)}\}_\itn$ and $\{\vy^{f(i)}\}_\itn$, number of epochs $n_e$,
		ANN $\nlinpart$}, Kalman gain
	\Statex
	\Statex \textbf{\textit{Initialization}}:\;
	\State Set the initial hyper-parameters $\weight^0$
	\State Generate $N \times M$-size augmented dataset \Comment Eq. (\ref{eq:augdata})
	\State Compute the testing metric $m_0 =\widehat{\mse}^{\textnormal{vr}} (\weight^0| \mathcal{D}_S) $
	\Statex \Comment similar to Eq. (\ref{eq:MC_reduced_variance}) with test dataset
	\Statex
	\Statex \textbf{\textit{Training}}:\;
	\For{$\nu= 1, 2, \dots, n_e$}
		\State Train ANN using mini-batch stochastic gradient descent method~\cite{ kingma2014adam,amari1993backpropagation} to update $\weight^\nu$
		%\Statex
		\State Compute the testing metric: $m_\nu =\widehat{\mse}^{\textnormal{vr}} (\weight^\nu| \mathcal{D}_S) $
		%\Commen Eq.~(\ref{eq:model_slection_metric_ann}).
		\Statex 	\Comment similar to Eq. (\ref{eq:MC_reduced_variance}) with test dataset
		\Statex
		\State \textbf{if} {$m_\nu < \min \{m_0,m_1, \dots, m_{\nu -1}\} $} \textbf{then}
			\State $\quad$ $\weight = \weight^\nu$ \;\Comment Call-back procedure
	\EndFor
	\Statex
	\State \textbf{\textit{Model selection}} \Comment Eq. (\ref{eq:model_selection})
	\Statex
	\Statex
	\Return {Hyper-parameters $\weight$ and vector $a$}
	\end{algorithmic}
\end{algorithm}

%%%%%%%%%%%%%%%%%%%%%%%%%%%%%%%%%%%%%%%%%%%%%%%%%%
\section{Numerical experiments: results and discussion}\label{sec:numerical_examples}
In this section, we demonstrate the performance of ML-EnCMF for the data assimilation of Lorenz 63 (L63)
and Lorenz 96 (L96) systems in comparison with the EnKF and LL-EnCMF.
Moreover the challenge in applying the ML-EnCMF for large-scale data assimilation applications is also discussed.
This section is organized as follows.
In Sec.~\ref{sec:numerical_examples:setup}, we first describe the general setup of the experiments.
Then, in Sec.~\ref{sec:lorenz63} and~\ref{sec:numerical_results:lorenz96},
	we discuss the numerical results for tracking L63 and L96 systems, respectively.
Finally, in the Sec.~\ref{sec:numerical_results:discussion}
 we discuss the computational cost of the ML-EnCMF and the challenges for
  large-scale data assimilation applications.

\subsection{Setup}\label{sec:numerical_examples:setup}

This subsection describes the setup of the investigated data assimilation problems, particularly dynamical models, %, observation scenarios,
performance metrics, and a numerical experiment for assessing errors in approximating the CM.
\subsubsection{Dynamical models}
\subsubsection*{Lorenz-63 system}
The L63 model is a simplified model of atmospheric convection \cite{lorenz1963deterministic} comprising three ODEs:
\begin{equation}
\dfrac{\dint \cq_1}{\dint t} = \const{\sigma} (\cq_2-\cq_1),
\quad \dfrac{\dint \cq_2}{\dint t} = \cq_1(\const{\rho} - \cq_3) - \cq_2,
\quad\dfrac{\dint \cq_2}{\dint t} = \cq_1\cq_2 - \const{\beta} \cq_3,
\end{equation}
where $\cq_1$, $\cq_2$, and $\cq_3$ are proportional to the rate of convection,
horizontal temperature variation, and vertical temperature variation, respectively,
and $\const{\sigma}$, $\const{\beta}$, and $\const{\rho}$ are the system parameters.
We use $\sigma= 10$, $\beta = 8/3$, and $\rho= 28$ as a conventional setting of the L63 system, resulting in a chaotic behavior.

\subsubsection*{Lorenz-96 system} The L96 model is an idealized model of a one-dimensional latitude band of
the Earth's atmosphere~\cite{lorenz1996predictability}.
The system is defined using a set of ODEs over the periodic domain of a 40-dimensional state vector,
$\cvq = [\cq_1,\cq_2, \dots, \cq_{n}]^{\top}$ with $n = 40$:
\begin{equation}
\begin{aligned}
\dfrac{\dint \cq_{\alpha}}{dt} &= (\cq_{\alpha+1} -
\cq_{\alpha-2})\cq_{\alpha-1} - \cq_\alpha + \const{F}, \quad  \alpha= 1,\dots, 40;\;
\cq_{n+1} = \cq_{1},
\end{aligned}
\end{equation}
where $\cq_0 = \cq_n$, $\cq_{-1} = \cq_{n-1}$, and $\const{F}$ is the forcing constant.
We select $\const{F}=8$, which is known to cause a chaotic behavior.

For both models, the \textit{synthesized} ground-truth state vector at $t=0$, $\vqtrue(t=0)$,
is given as a sample of the normal distribution $\mathcal{N}(\cvzero_{n},\cmI_{n})$.
The {\rv} $\rvQ_{0}$ is also assumed to follow this distribution.
We simulate the L63 and L96 models using a fourth-order explicit Runge-Kutta algorithm with a time step $\Delta t = 0.01$.

\subsubsection{Performance metrics}
At each assimilation step $k$, the mean vector $\overline{\vq}^a_k$ of the updated ensemble
$\{\vq^{a (i)}_k\}_\itn$ is compared with the ground-truth state $\vqtrue(t_k)$ using the component-average root MSE (RMSE) criterion, $rmse_k$,
 which is defined as
\begin{equation}
{rmse}_k=\dfrac{\lVert\overline{\vq}^a_k- \vqtrue(t_k)\rVert}{\sqrt{n}},
\quad k=1, 2, \dots, \kappa,
\end{equation}
where $n=3$ for L63 and $n=40$ for  L96.
We use the
\emph{average} of these values  $\{{rmse}_1,\; {rmse}_2, \dots, {rmse}_{\kappa} \}$ as a performance metric.
We also monitor the average ensemble spread, $\overline{ens}$, defined as 
\begin{equation}\label{eq:ens}
\overline{ens} =\dfrac{1}{\kappa}\sum_{k=1}^\kappa \left [\dfrac{\tr ({\cov{\rvQ^a_k}})}{n} \right ]^{1/2},
\end{equation}
to measure the contraction of the ensemble. In Eq.~(\ref{eq:ens}),
$\tr(\cov{\rvQ^a_k})$ is estimated from ensemble $\{\vq^{a (1)}_k, \dots,\vq^{a (N)}_k \}$.
The final metric that we consider herein is the average coverage probability $f_{\textnormal{cv}}$
 of the $95\%$-confidence interval bounded between $2.5\%$ and $97.5\%$ quantiles of each marginal distribution,
  which is estimated as follows
\begin{equation}
f_{\textnormal{cv}} = \dfrac{1}{n \times \kappa} \sum_{k=1, 2, \dots,\kappa;\; \alpha = 1,2, \dots,n}
\mathbf{1}(\vqtrue_{\alpha}(t_k)\in {I_{\alpha,k}}),
\end{equation}
where
$I_{\alpha,k}$ is  the estimated $95\%$-confidence interval of the $\alpha$-{th} component of the updated ensemble at $t_k$.
For small ensemble sizes,
 \emph{e.g.}, $N= 20$, 30, and 60, we compute the $90\%$, $93.3\%$, and $93.3\%$-coverage probability, respectively.
The aforementioned performance metrics are commonly used to evaluate data assimilation algorithms, see \emph{e.g.}
~\cite{lei2011moment, Spantini2019, lee2016state}.

\subsubsection{Assessing errors of conditional mean approximations}\label{sec:numerical_experiment}
Here we discuss a numerical experiment for assessing errors in approximating the CM
 using the ML-EnCMF and LL-EnCMF caused by small ensemble sizes $N$.
First, a filtering procedure with $\kappa$ assimilation steps using LL-EnCMF
with a large ensemble size
 $N_{\text{ref}}$, \emph{i.e.}, $N_{\text{ref}}\gg N$,
 is implemented as the reference EnCMF.
 The reference EnCMF provides an approximation of the CM
  at each assimilation step with negligible errors.
Secondly, the forecast ensemble of the state and measurement at each assimilation step $k=1,\dots, \kappa$,
 $\{(q^{f(i)}_k, y^{f(i)}_k)\}_{1}^{N_{\text{ref}}}$,
 is randomly divided into two sets $D_1$ and $D_2$ with sizes $N$ and $N_{\text{ref}}-N$, respectively.
We then apply the ML-EnCMF and LL-EnCMF to approximate the CM using
set $D_1$ as the input forecast ensemble
and compute their component-average RMSE $rmse_{\phi}$ based on set $D_2$ as
\begin{equation}
rmse_{\phi,k} = \left[\dfrac{1}{{N_{\text{ref}}-N}}\sum_{(x^f, y^f) \in D_2 }
\dfrac{\Vert \widehat{\phi}_{\rvQ^f, k} (y^f|D_1) - \phi_{\rvQ^f, k}^{
\text{ref}} (y^f) \Vert^2}{n} \right ]^{1/2}, \quad k = 1,\dots, \kappa,
\end{equation}
where $\widehat{\phi}_{\rvQ^f, k}(\cdot|D_1)$ is the approximation of the CM map $\phi_{\rvQ^f, k}$
via the ML-EnCMF (Eq.~(\ref{eq:cm_approximation})) or the LL-EnCMF (Eq.~\ref{eq:ll_cm_next}) using
the same $N$-sized forecast ensemble $D_1$,
and $\phi_{\rvQ^f, k}^{\text{ref}}$ is the reference solution.
Finally, we evaluate the average value of $rmse_{\phi,k}$ as
\begin{equation}
\overline{rmse}_{\phi} = \dfrac{1}{\kappa} \sum_{k=1}^{\kappa} rmse_{\phi,k}.
\end{equation}
We use the average value $\overline{rmse}_{\phi}$ as an indicator for the misfit between the reference CM and its approximations.
Running this numerical experiment requires a significant computational cost due to the condition $N_{\text{ref}}\gg N$.
Therefore, we implement the experiment only for L63 cases.

\subsection{Lorenz 63 system}\label{sec:lorenz63}
This subsection reports the numerical results in tracking the L63 system.
The data assimilation setting is described as follows.
The initial ensembles are obtained by running the EnKF for 2000 assimilation steps.
Then, we apply different techniques, \emph{i.e.}, EnKF, ML-EnCMF, LL-EnCMF,
for the next 2000 steps and evaluate their performance.
We assume that every state is observed using the complete observation model:
\begin{equation}%\label{eq:linear_observation}
\yob_{k} =\vqtrue(t_k) + \vxi_{k},\quad \varXi_{k}\sim \mathcal{N}(\cvzero_3,\; 2^2\;\cmI_3).
\end{equation}
Here we consider two scenarios for the observational period $\Delta T_{\mathrm{obs}} := t_{k+1} - t_k$,  $\Delta T_{\mathrm{obs}}=0.5$ and $\Delta T_{\mathrm{obs}}=1$.
The selected observational periods are more extended than the usually reported ones ($\Delta T_{\mathrm{obs}}\leq 0.1$),
which amplifies the nonlinear characteristic.

We implement the ML-EnCMF using a two-hidden-layer dense ANN structure with 20 nodes at each hidden layer
and the rectified linear unit as the activation function.
The variance reduction technique, see Eq.~(\ref{eq:MC_reduced_variance}), is applied such that $N \times M=6000$.
The forecast ensembles are divided into training and testing datasets with a size ratio of $N_T:N_S = 8:2$.
At each assimilation step, we train the ANN for a maximum of 100 epochs, a learning rate of 0.001,
and a batch size of 128 using Algorithm~\ref{algorithm:meta_training}.

For the LL-EnCMF,
we use inflation to avoid degeneracy.
The inflation modifies the updated ensemble as
\begin{equation}\label{eq:inflation}
	\vq^{a(i)} \leftarrow \overline{\vq}^{a}+ \delta (\vq^{a(i)} - \overline{\vq}^{a}), \quad i =1, \dots, N,
\end{equation}
where $\overline{\vq}^{a}$ is the mean of the ensemble, and $\delta$ ($\geq 1$)  is the inflation coefficient.
For a sufficiently large ensemble size, $\delta = 1$, in other words, the inflation is not active.
We test different inflation coefficients in range $[1, 1.3] $ and select those yielding the best results.
We note that the inflation technique is not applied for the ML-EnCMF or EnKF as an additional challenge
 for assessing these methods' robustness.

We run four data assimilation procedures, which perform $\kappa =2000$
updating steps each following the above-described setting.
The estimated performance metrics vary insignificantly when
adding more assimilation procedures.
The average performance metrics of the
ML-EnCMF, EnKF, and LL-EnCMF for $\Delta T_{\mathrm{obs}}=0.5$ and $\Delta T_{\mathrm{obs}}=1$ are reported
on Tab.~\ref{tab:l63_05} and~\ref{tab:l63_1}, respectively.
Here, one can compare the average RMSE of the updated mean
with the standard deviation of the measurement error
to obtain the relative magnitude.

\begin{table}[h!]
\centering
	\caption{L63 system: performance metrics (average RMSE - average spread (average coverage probability $f_{\textnormal{cv}}$))
		of the ML-EnCMF compared with the
	LL-EnCMF and EnKF  with $\Delta T_{\mathrm{obs}}=0.5$.
		The LL-EnCMF uses inflation coefficients $1.25$, $1.2$, and $1.05$ for ensemble sizes 20, 30, and 60, respectively.
		For $N=20$, 30, and 60 we compute the coverage probability
     for 90\%, 93.3\%, and 93.3\%-confident intervals, respectively.}
	\begin{tabular}{|l|c|c|c|c|c|}
		\hline
		\hline
		$N$		& 20				& 30 				& 60 				& 100 				& 200\\
		\hline
		ML-EnCMF&1.27-0.96			& 1.11 - 0.97 		& 0.94 - 0.97 		& 0.86 - 0.97 		& 0.81 - 0.96  \\%done 10.01.2021
				&(0.83)				& (0.90)  			& (0.92) 	  		& (0.94) 			& (0.95)\\
		\hline
		EnKF 	&1.37-1.17			& 1.29 - 1.21 		& 1.24 - 1.27 		& 1.23 - 1.29 		& 1.22 - 1.30  \\%done 10.01.2021
		     	&(0.86)				& (0.90)      		& (0.93)      		& (0.93)  			& (0.93)    \\
		\hline
		%LL-EnCMF& 1.38-0.80  		& 1.17-0.87 		& 0.85- 0.90 		& 0.81 - 0.90  \\%done 10.01.2021
		%		& (0.78)       		& (0.89)      		& (0.93)  			&  (0.95) \\ delta = 1.02
		LL-EnCMF&1.43 - 1.02		& 1.21-1.04  		& 0.99-0.91 		& 0.90-0.89			& 0.85 - 0.90  \\%done 10.01.2021
				&(0.85)				& (0.92)       		& (0.93)      		& (0.93)  			& (0.95) \\
		\hline
	\end{tabular}
	\label{tab:l63_05}
\end{table}

\begin{table}[h!]
\centering
	\caption{L63 system: performance metrics (average RME - median RMSE (average coverage probability $f_{\textnormal{cv}}$))
		of the ML-EnCMF compared with the
	LL-EnCMF and EnKF  with $\Delta T_{\mathrm{obs}}=1$.
			The LL-EnCMF uses inflation coefficients $1.25$, $1.2$, and  $1.1$ for ensemble sizes 20, 30, and 60, respectively.
			For $N=20$, 30, and 60 we compute the coverage probability
     for 90\%, 93.3\%, and 93.3\%-confident intervals, respectively.}
	\begin{tabular}{|l|c|c|c|c|c|}
		\hline
		\hline
		$N$ 	 &20			&30 				& 60 			    & 100 				& 200 \\
		\hline
		ML-EnCMF &1.66 - 1.19	& 1.50 - 1.20 		& 1.22 - 1.21	    & 1.14-1.18 		& 1.06 - 1.17 \\ %done 10.01.2021
				 &(0.82)		& (0.90) 	 		& (0.90)    	    & (0.93) 			& (0.93) \\ %done 10.01.2021
		\hline
		EnKF 	&1.67 -1.46		& 1.61 - 1.55 		& 1.53 - 1.57 	    & 1.51 - 1.60 		& 1.51 - 1.61 \\ %done 10.01.2021
			 	&(0.87)			& (0.92) 			& (0.91)		    & (0.93) 			&  (0.93)\\ %done 10.01.2021
		\hline
		LL-EnCMF&2.52 - 1.11	& 1.78-1.15  		& 1.35 - 1.12 		& 1.18 - 1.06 			& 1.05 - 1.06  \\%done 10.01.2021
				&(0.72)			& (0.86)       		& (0.90)      	    & (0.93)  			&  (0.93) \\
		\hline
	\end{tabular}
	\label{tab:l63_1}
\end{table}

In comparison with the EnKF, the ML-EnCMF consistently exhibits better performance.
The improvement increases with the ensemble size because the nonlinearity is better approximated.
For instance, with $N=200$, the average RMSE is reduced by 34\% and 30\%, for $T_{\mathrm{obs}}=0.5$
and $T_{\mathrm{obs}}=1$,  respectively.
A similar trend is also observed for the average ensemble spread.
While the average ensemble spread is significantly reduced using the ML-EnCMF,
the average coverage probability is still comparable with the EnKF.
It is noteworthy that, the proposed filter shows better performance not only with large ensemble sizes but also
small ones, \emph{e.g.}, $N\leq 60$.
For $N=30$ and $T_{\mathrm{obs}}=1$,the EnKF and ML-EnCMF show similar performance
because the ANN cannot capture the CM's nonlinearity with
a small ensemble size and under significant uncertainty
(compared with the case of $N=30$ and $T_{\mathrm{obs}}=0.5$)

Compared with the LL-EnCMF, the ML-EnCMF exhibits significantly better performance
in terms of average RMSE for small ensemble sizes $N\leq 60$,
which is a crucial advantage of the ML-EnCMF because the ensemble size is usually small in practice.
Notably, the LL-EnCMF exhibits worse performance compared with the EnKF for $N=20$ and $N=30$ ($T_{\mathrm{obs}}=1$).
Another disadvantage of the LL-EnCMF compared to the ML-EnCMF is the strong dependence on the inflation coefficient.
For example, the average RMSEs obtained by the LL-EnCMF
without using the inflation technique in case $\Delta T_{\mathrm{obs}}=0.5$ are  2.33 and 1.65,
for ensemble sizes of 20 %rmse: 1.98,2.39, 2.67, 2.22
and 30, %rmse: 2.07,1.5,1.43,1.5
%and 60, %rmse: (0.96, 0.98, 1.03, 1.04)
respectively.
For $N\geq 100$, both methods exhibit similar performance metrics.

\begin{figure}[h!]
	\centering
	\begin{subfigure}[b]{0.48\textwidth}
	    \centering
		\includegraphics[scale = 0.4]{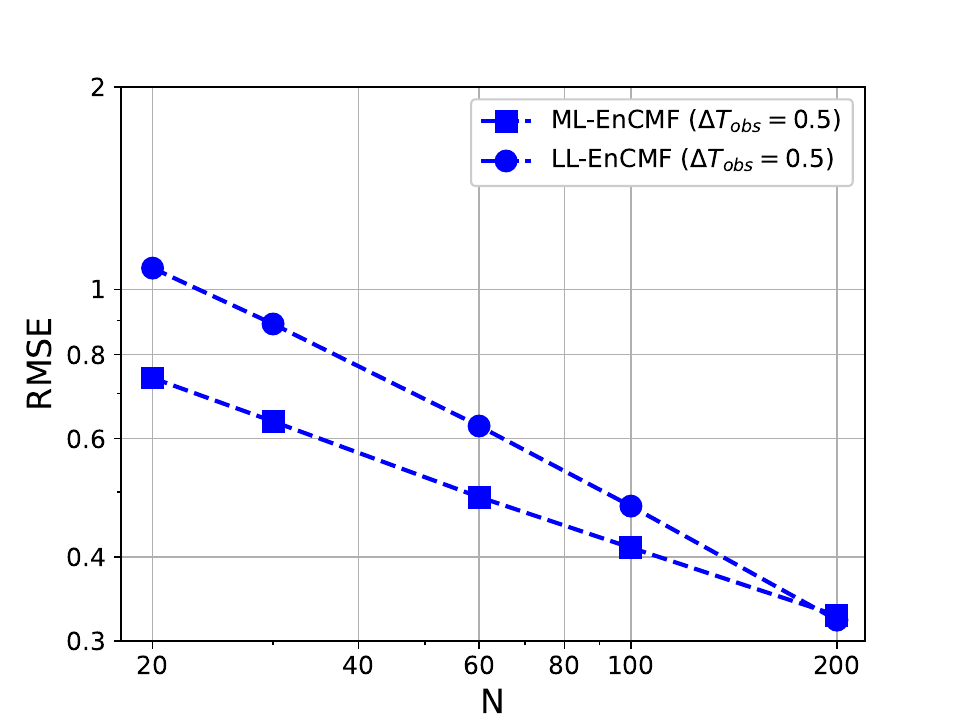}
		\caption{$~$}
	\end{subfigure}
	\begin{subfigure}[b]{0.48\textwidth}
	    \centering
		\includegraphics[scale = 0.4]{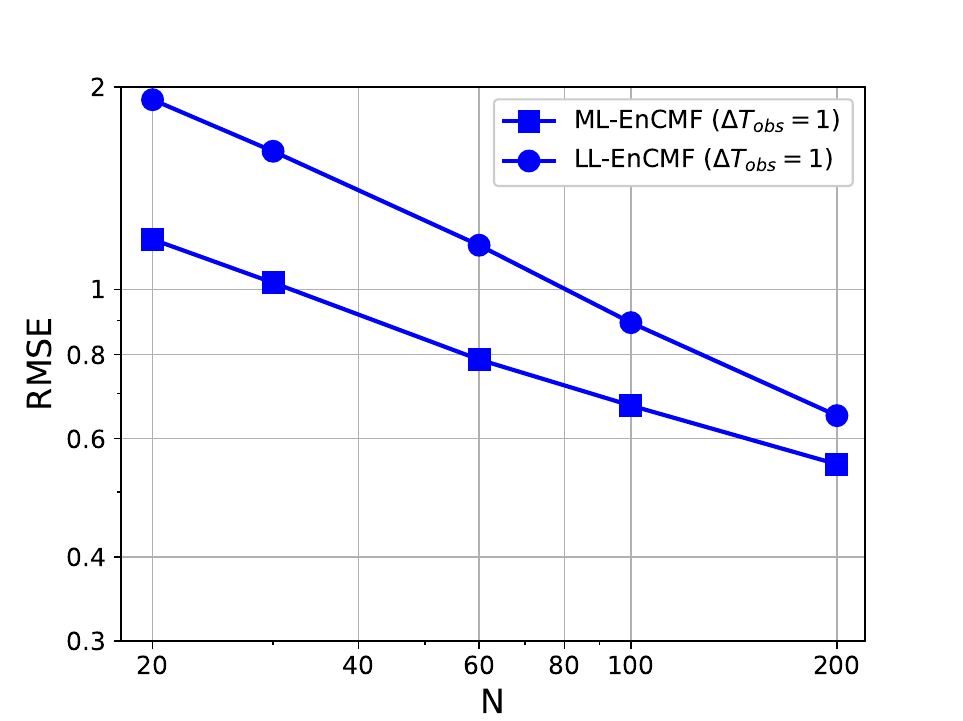}
		\caption{$~$}
	\end{subfigure}
	\caption{Lorenz 63: average RMSE of the CM approximations
	using LL-EnCMF and ML-EnCMF $\overline{rmse}_{\phi}$, a) $\Delta T_{\text{obs}} =0.5$,
	b) $\Delta T_{\text{obs}} = 1$.}
	\label{fig:rmse_phi}
\end{figure}

We implement the numerical experiment discussed in Sec.~\ref{sec:numerical_experiment}
using $N_{\text{ref}}= 10000$.
 The average RMSEs of the CM approximations using the ML-EnCMF and LL-EnCMF $\overline{rmse}_{\phi}$
  are plotted in Fig.~\ref{fig:rmse_phi}.
%For $\Delta T_{\mathrm{obs}}=0.5$ and $N=200$, the ML-EnCMF exhibits a slightly bigger error than the LL-EnCMF.
% Beside that particular scenario,
The CM approximating errors of ML-EnCMF  are smaller compared with the LL-EnCMF for
$N\leq 60$.
Furthermore, those errors of ML-EnCMF and LL-EnCMF are getting closer as the ensemble size increases.
For $N \geq 100$, the approximating errors yielded by the ML-EnCMF and LL-EnCMF
are negligible compared with the standard deviation of the measurement error.
 This trend of the approximating CM errors explains the previous numerical observation
 that the ML-EnCMF outperforms the LL-EnCMF for small ensemble sizes ($N\leq 60$),
 while for $N\geq 100$ both methods exhibit similar performance metrics.
\subsection{Lorenz 96 system}\label{sec:numerical_results:lorenz96}
This subsection reports the numerical results in tracking the L96 system.
The initial ensembles are obtained by running the EnKF for 2000 assimilation steps using a complete observation model with standard Gaussian noise.
Then, we apply different techniques, \emph{i.e.}, EnKF, ML-EnCMF, and LL-EnCMF,
for the next 2000 steps using an incomplete observation model:
\begin{equation}\label{eq:incomplete_observation}
\yob_k = \mat{H} \vqtrue(t_k) + \vxi_k,\quad \rvXi_{k}\sim \mathcal{N}(\cvzero_{20},\; 0.5\;\cmI_{20}),
\end{equation}
where $\mat{H} \in \mathbb{R}^{m\times n}$ with $m = 20$
is a linear operator that selects the even-indexed components $\vqtrue_{2} (t_k), \vqtrue_{4} (t_k), \dots,\vqtrue_{40} (t_k)$ from $\vqtrue (t_{k})$.
We examine the observation time interval $\Delta T_{\textnormal{obs}}=0.4$.
For reference, $\Delta T_{\textnormal{obs}} = 0.05$ is comparable to $6$ h in a weather forecast model~\cite{lorenz1998optimal, majda2012filtering}. %\sidenote{05.02 done}
Due to the long observational period and the incomplete observation model,
the relation between an observed component and an unobserved one is highly nonlinear.
Moreover, the distribution of the forecast is no longer Gaussian.
The considered setting is hence referred to as \emph{the hard case} in~\cite{lei2011moment}.

The implemented ML-EnCMF in this example uses a one-hidden-layer ANN structure with 40 nodes in the middle layer
for approximating the CM.
Instead of a dense ANN structure, a \emph{localized} one is used.
There are two motivations for applying such structures:
i) to explicitly remove correlation between components with a long spatial range ---
a widely applied technique for the EnKF and LL-EnCMF,
ii) to reduce the number of hyperparameters required to be trained.
Without using the localized structure, the performance of the ML-EnCMF may not optimized for small ensemble sizes.
The ANN structure designed for tracking the L96 system is described as follows.
Let $u := g_{NN} (y)$,
where $g_{NN}: \sR^{20} \rightarrow \sR^{40}$ (see Eq.~(\ref{eq:cm_approximation})). The vector $u$ is evaluated as
\begin{equation}
\begin{aligned}
v_{\alpha} &= \sigma_1 \left(\sum_{\alpha'=2, 4, \dots, 40} w^1_{\alpha \alpha'}\; y_{\alpha'/2} \; \mathbf{1} (z_{\alpha \alpha'}\leq l_1)
+\gamma^1_{\alpha} \right)\;,\quad \alpha=1,\dots, 40\\
u_{\alpha} &= \sum_{\alpha'=1}^{40} w^2_{\alpha \alpha'}\;v_{\alpha'} \; \mathbf{1} (z_{\alpha \alpha'}\leq l_2) +\gamma^2_{\alpha}\;,\quad \alpha=1,\dots, 40
\end{aligned}
\end{equation}
where $z_{\alpha \alpha'} = \min(|\alpha-\alpha'|, 40-|\alpha-\alpha'|)$ is the spatial distance between components $\alpha$ and $\alpha'$ of the state vector,
$v_1, \dots, v_{40}$ are the output of the hidden layer,
$w^1_{\alpha \alpha'}$ ($w^2_{\alpha \alpha'}$) and $\gamma^1_{\alpha}$ ($\gamma^2_{\alpha}$) are weights and biases
of the hidden (output) layer, respectively,
$\sigma_1$ is the activation function of the hidden layers,
and $l_1$ and $l_2$ are the \emph{localization lengths}.
In a dense version, $l_1$ and $l_2$ are equal to 20.
Here we choose $l_1=l_2=3$ to incorporate sparsity.
The number of trainable weights is hence reduced to 420 from 2400 in the dense version.
For training the ANN, the variance reduction technique, see Eq.~(\ref{eq:MC_reduced_variance}), is applied such that $N \times M \simeq 10000$.
The other training settings are as described in the L63 example above.

The localization approach is also applied for the EnKF, the linear part of the ML-EnCMF, and the LL-EnCMF.
For implementing the EnKF and the linear part in the ML-EnCMF, see Eq.~(\ref{eq:cm_approximation}),
we use the Cohn-Gaspari covariance tapering~\cite{gaspari1999construction} with $c=20$ for $N\leq 300$ and
$c=30$ for $N=400$.
For the LL-EnCMF, we use the localization technique described in~\cite{lei2011moment}.

The inflation technique is applied for all tested filters in this \emph{hard case} example.
The selected inflation coefficient for the EnKF is $\delta=1.005$.
For the ML-EnCMF and LL-EnCMF, we test different inflation coefficients in ranges $[1., 1.05]$ and $[1., 1.3]$, respectively,
and choose the best results.
The selected inflation coefficients for the ML-EnCMF are 1.05 ($N=100$), 1.03 ($N\in \{100, 200\}$), 1.01 ($N=300$), and 1.005 ($N=400$).
Those of the LL-EnCMF are 1.2 ($N=100$), 1.18 ($N\in \{150, 200\}$), and 1.1 ($N\in\{300, 400\}$).%. and 1.06 ($N=600$).
%Other settings of the LL-EnCMF are as discribed in~\cite{lei2011moment}.

We run four data assimilation procedures with different generating seeds.
The average performance metrics are reported on Tab~\ref{tab:l96_04}.
For cross-reference, in~\cite{lei2011moment}, the average RMSEs obtained using the LL-EnCMF and EnKF with $N=400$
are 0.68 and 0.79, respectively.
These values are slightly different from our EnKF and LL-EnCMF results due to the non-deterministic of
the synthesized measurement data and statistical errors when evaluating the average values.
The LL-EnCMF performances for smaller ensemble sizes ($N<400$) are not reported in~\cite{lei2011moment}.
\begin{table}[h!]
\centering
	\caption{L96 system: performance metrics, \emph{i.e.}, average RMSE - average spread (average coverage probability $f_{\textnormal{cv}}$),
		of the ML-EnCMF compared with the
	LL-EnCMF and EnKF  with $\Delta T_{\mathrm{obs}}=0.4$.}
	\begin{tabular}{|l|c|c|c|c|c|}
		\hline
		\hline
		$N$		&100				& 150 				& 200 				& 300 					& 400 				\\
		\hline
		ML-EnCMF&0.84 -	0.70		&0.79 - 0.67		& 0.75 - 0.67 		&0.72 - 0.64			& 0.69 - 0.63 		\\
				&(0.94)				& (0.94)  			& (0.94) 	  		&(0.94)					& (0.94) 			\\
				%&1.05(done)		&(1.03)done			&(1.03)done			&(1.01)done				& (1.005)			\\
		\hline
		EnKF 	&0.88 - 0.71		&0.85 - 0.73		& 0.83 - 0.74 		&0.84 - 0.76			& 0.83 - 0.78 		\\
		     	&(0.92)				& (0.93)    		& (0.94)    		&(0.94) 				& (0.95)  			\\
		\hline
		LL-EnCMF&1.21 - 0.74		&0.92 - 0.73		& 0.88 - 0.76		&0.74 - 0.68			& 0.70 - 0.69 		\\
				&(0.89)				& (0.93)  			& (0.95) 	  		&(0.95)					&(0.96) 			\\
		\hline
	\end{tabular}
	\label{tab:l96_04}
\end{table}

We observe a pattern similar to the L63 system.
Indeed, the ML-EnCMF consistently outperforms the EnKF in terms of average RMSE and average spread
thanks to its ability to approximate the nonlinearity of the CM.
In terms of 95\%-coverage probability, the ML-EnCMF and EnKF have comparable performance.

Compared with the LL-EnCMF, the ML-EnCMF and LL-EnCMF exhibit comparable performance for  $N = 400$.
However, the ML-EnCMF outperforms the LL-EnCMF for smaller ensemble sizes  due to the significant statistical error when approximating the CM via
the likelihood function with a small ensemble size.
Notably, the LL-EnCMF shows worse performance compared to the EnKF for $N\leq 200$. %While in our approach, the model selection (Sec.~\ref{sec:cross_validation})

\subsection{Discussion on large-scale applications}\label{sec:numerical_results:discussion}
The ML-EnCMF practicality requires that the computational time for
 ANN training and other computational tasks is shorter than the prediction range.
Data assimilation applications have varying temporal prediction ranges in practice.
For example, the prediction ranges of weather and climate applications
 can be classified into short-range (one to two days), medium-range (days to weeks), and (sub)seasonal-range.

In our numerical experiments, the typical training time for approximating the CM is 22-27 (s) and 32-37 (s) for L63 with an ensemble size of 200 and L96 with an ensemble size of 400, respectively, on one core of a 2.8 GHz Quad-Core Intel Core i7 CPU.
 The training time in the L96 case does not significantly grow owing to the sparse ANN structure and the applied early stopping.
  Furthermore, the sizes of augmented datasets are identical for the L63 and L96 experiments.

We project that the ML-EnCMF fulfills the computational time constraint
 for applications with moderately high-dimensional states with a short (or longer) prediction range and
for larger systems with medium to seasonal ranges. Moreover, the ANN training process is accelerated by
 using advanced hardware, \emph{e.g.}, GPU, as well as improving the algorithm,
  \emph{e.g.}, the ANN can be pre-trained to learn the manifold of the states
  to reduce the online-training time on that ANN~\cite{lin2008riemannian, soize2021probabilistic}.
In~\cite{chantry2021opportunities}, Chantry \emph{et al.} discuss in more detail the opportunities
and challenges for machine learning in weather and climate modeling.
 We will investigate applications of the ML-EnCMF for larger systems with the vector-field state in our future research.

%%%%%%%%%%%%%%%%%%%%%%%%%%%%%%%%%%%%%%%%%%%%%%%%%%
\section{Conclusion}\label{sec:conlusion}
This work analyzes the properties of the CMF and develops the ML-based ensemble method for its implementation.
The CMF's updated mean matches that of the posterior, obtained by applying Bayes' rule on the filter's forecast distribution.
Moreover, the CMF's updated covariance coincides with the expected conditional covariance.

We develop ML-EnCMF using ANNs and based on the orthogonal projection property of the CM.
The filter naturally inherits the robust performance of the EnKF in closely linear Gaussian scenarios,
 owing to the combination of ANNs with the EnKF's updating map.
A systematical methodology for integrating machine learning approaches into the EnCMF is developed.
We apply a variance reduction technique to augment the dataset and reduce statistical errors.
Moreover, we perform a model selection procedure at each updating step for selecting element-wisely the applied filter,
 \emph{i.e.}, either the EnKF or the ML-EnCMF.
Notably, the model selection ensures that our filter outperforms the EnKF.

We demonstrate the ML-EnCMF performance using the Lorenz-63 and Lorenz-96 systems.
Notably, for the Lorenz-96 system,
we propose a localized ANN structure to avoid overestimated long-range correlation
and reduce computational costs for ANN training.
In summary, the proposed filtering method exhibits considerable improvement compared with the commonly used methods,
such as the EnKF and the LL-EnCMF, particularly for small ensemble sizes.

For future work, the following developments will be considered:
i) developing the ML-EnCMF for large-scale systems with vector-field state,
ii) enhancing the computational efficiency of the ML-EnCMF and
iii) combining the filter with the method of A-optimal design of experiments.

%%%%%%%%%%%%%%%%%%%%%%%%%%%%%%%%%%%%%%%%%%%%%%%%%%
\section*{Acknowledgment}
%\rjp{We would like to thank the anonymous reviewers for helpful suggestions that significantly improved this work.} 
This publication was supported by funding from the Alexander von Humboldt Foundation and
King Abdullah University of Science and Technology (KAUST) Office of Sponsored Research (OSR) under award numbers\\
URF/1/2281-01-01 and URF/1/2584-01-01 in the KAUST Competitive Research Grants Programs, respectively.
R. Tempone is a member of the KAUST SRI Center for Uncertainty Quantification in Computational Science
and Engineering.
Simulations were performed with computing resources granted by RWTH Aachen University under project rwth0632.

%%%%%%%%%%%%%%%%%%%%%%%%%%%%%%%%%%%%%%%%%%%%%%%
\appendix

%%%%%%%%%%%%%%%%%%%%%%%%%%%%%%%%%%%%%%%%%%%%%%%%%%
\section{Kalman filters}\label{appendix:enkf}
This appendix summarizes the Kalman filter and its ensemble version.
\subsection{Kalman filter}
In the linear-Gaussian setting, the {\rv s} $\rvQ^f_k$ and  $\rvY^f_k$ are considered Gaussian.
Consequently, the conditional PDF stated in Eq.~\eqref{eq:bayesian_posterior_PDF} is simplified to a normal
distribution PDF.
Let $h(\vq) = \mat{H}\vq$, where $\mat{H} \in \sR^{m\times n}$.
The closed-form of the {\rv} $\rvQ^{a}_k$, whose probability PDF is identical to that of the local
posterior~\cite{Kalman1960}, is expressed as follows:
\begin{subequations}
	\begin{align}
&\rvQ^{a}_k  = \rvQ^f_k + \mat{K}_k^l(\yob_k -\rvY^f_k),\label{eq:KF_filter_interms_rv}\\
& \mat{K}_k^l = \mat{\varSigma}_{\rvQ_k^f}\mat{H}^{\top}(\mat{\varSigma}_{\rvXi_k}+
\mat{H}\mat{\varSigma}_{\rvQ_k^f}\mat{H}^{\top})^{-1},%\mat{K}_k= \cov{\rvQ^f_k,\rvY^f_k}\variance{\rvY^f_k}^{-1},
\label{eq:kalman_simple_form}
\end{align}
\end{subequations}
where $\rvQ^f_k$ is evaluated using the model ${\dmodel}_k$ assumed to be linear, $\mat{K}^l_k$ is the Kalman gain,
and $\mat{\varSigma}_{\rvQ_k^f}$ and $\mat{\varSigma}_{\rvXi_k}$ are the covariance matrices of the {\rv s}
$\rvQ_k^f$ and $\rvXi_k$, respectively.
 The transformation in Eq.~\eqref{eq:KF_filter_interms_rv} is a linear version of the map $\mathcal{T}_k$
(Eq.\eqref{eq:general_form_filter}).
In the linear-Gaussian setting, the {\rv} $\rvQ^a$ obtained using Eq.~\eqref{eq:KF_filter_interms_rv} is a Gaussian
{\rv}, hence, fully characterized {by} its mean and covariance.
From Eq.~\eqref{eq:KF_filter_interms_rv}, the formulations for updating these statistical moments can be
straightforwardly derived, {known as the KF}~\cite{Kalman1960}.

\subsection{Ensemble Kalman filter}
The EnKF is an ensemble implementation of the KF for dealing with nonlinear dynamical systems, in which the forecast
and updated {\rv s} are represented using the ensembles of their samples, thereby allowing the approximation of
non-Gaussian distributions~\cite{Evensen2009}.
However, the EnKF applicability still requires a linear observation map.
Let $\{\vq_{k-1}^{a(i)}\}_\itn$ be the $N$-sample ensemble of the {\rv} $\rvQ^a_{k-1}$.
The ensembles  $\{\vq_{k}^{f(i)}\}_\itn$ and $\{\vy_{k}^{f(i)}\}_\itn$ of the {\rv s}
$\rvQ^f_{k}$ and $\rvY^f_{k}$, respectively, are obtained using Eq.~(\ref{eq:forecast}).
For the update step, we first estimate the mean $\overline{\vq}_k^f$ and the covariance matrix
$\bm{\varSigma}_{\rvQ^f_k}$ of the {\rv } $\rvQ^f$ using the forecast ensembles:
\begin{equation}\label{eq:forecast_moment_MC}
	\begin{aligned}
		\overline{\vq}_k^f &\approx \dfrac{1}{N} \sum_{i=1}^N \vq_k^{f(i)},\\
		\bm{\varSigma}_{\rvQ^f_k} &\approx \dfrac{1}{N} \sum_{i=1}^N \left ( \vq_k^{f(i)} - \overline{\vq}_k^f\right ) \left (\vq_k^{f(i)} - \overline{\vq}_k^f \right )^{\top}.
	\end{aligned}
\end{equation}
The obtained covariance matrix is then used to compute the Kalman gain by evaluating Eq.~(\ref{eq:kalman_simple_form}).
Subsequently, the EnKF uses the linear update map of the KF, as expressed in Eq.~\eqref{eq:KF_filter_interms_rv},
and employs the ensemble technique to approximate the updated ensemble:
\begin{equation}\label{eq:enkf}
 \vq^{a(i)}_k = \vq^{f(i)}_k + \mat{K}_k^l(\yob_k -\vy_k^{f (i)}), \quad i= 1,\dots, N,
\end{equation}
where $\{\vq_{k}^{a(1)}, \dots, \vq_k^{a(N)}\}$ is the updated ensemble representing the {\rv} $\rvQ^a_k$.

\section{Likelihood-based approximation of the CM and the LL-EnCMF}\label{appendix:likelihood_approach}
Combining Eqs.~(\ref{eq:bayesian_posterior_PDF_global}) and (\ref{eq:conditional_mean}),
the conditional mean $\phi_{\rvQ^f}$ can be evaluated via the likelihood function as
\begin{equation}\label{eq:ll_cm}
\phi_{\rvQ^f}(\vy) = \int_{\sR^n} \vq \dfrac{\pi_{\rvQ^f}(\vq) \; \pi_{\rvXi} (\vy - h(\vq))}
{\int_{\sR^n} {\pi_{\rvQ^f}(\vq') \; \pi_{\rvXi}(\vy - h(\vq'))} \;\dint \vq'} \dint \vq.
\end{equation}
Similarly the conditional expectation of the second moment $\expectation{\left(Q^{f} \right)^2|Y^f}$ can be expressed as
\begin{equation}\label{eq:ll_cm2}
\expectation{\left(Q^{f} \right)^2|Y^f=\vy}= \int_{\sR^n} \vq^2 \dfrac{\pi_{\rvQ^f}(\vq) \; \pi_{\rvXi} (\vy - h(\vq))}
{\int_{\sR^n} {\pi_{\rvQ^f}(\vq') \; \pi_{\rvXi}(\vy - h(\vq'))} \;\dint \vq'} \dint \vq.
\end{equation}

We recall from Sec.~\ref{sec:cmf} that $\{\vq^{f(i)}\}_\itn$ and $\{\vy^{f(i)}\}_\itn$ are the forecast ensembles of the state and observation, respectively.
Implementing the EnCMF, see Eq.~(\ref{eq:en_cmf}), requires evaluating map $\phi_{\rvQ^f}$ for $\{\vy^{f(i)}\}_\itn$ and $\yob$.
Using the likelihood-based approach, the values of $\phi_{\rvQ^f}(\vy^{*})$ with $\vy^{*} \in \{\vy^{f(1)},\dots \vy^{f(n)},\yob\}$
are respectively approximated as
\begin{equation}\label{eq:ll_cm_next}
\widehat{\phi_{\rvQ^f}}(\vy^{*})=\dfrac{1}{N}\sum_{i=1}^N \vq^{f(i)} \dfrac{\pi_{\rvXi} (\vy^{*} - h(\vq^{f(i)}))}
{\dfrac{1}{N}\sum_{i=1}^N \; \pi_{\rvXi}(\vy^{*} - h(\vq^{f(i)}))},
	\quad \vy^{*} \in \{\vy^{f(1)},\dots \vy^{f(n)},\yob\},
\end{equation}
which is the ensemble-based version of the Eq.~(\ref{eq:ll_cm}).
The updated ensemble of the LL-EnCMF is then obtained as
\begin{equation}\label{eq:ll_encmf}
\vq^{a(i)} = \vq^{f(i)} + \widehat{\phi_{\rvQ^f}}(\yob) - \widehat{\phi_{\rvQ^f}}(\vy^{f(i)}) .
\end{equation}

%%%%%%%%%%%%%%%%%%%%%%%%%%%%%%%%%%%%%%%%%%%%%%%%%%
\section{Analysis of the variance reduced estimator}
\label{appendix:variance_duction}
To motivate that $\widehat{\mse}^{\textnormal{vr}}$  may provide an estimation of $\mse$ with reduced variance, we note that
\begin{equation}\label{eq:MC_augmenteddata}
\begin{aligned}
\lim_{M\rightarrow \infty}\widehat{\mse}^{\textnormal{vr}}(\weight | \mathcal{D}_T)
	&\;=_{\text{a.s.}}\;\dfrac{1}{N_T} \sum_{i=1,\dots, N_T}
	\mathbb{E} \Bigl[\bigl \lVert \vq^{f(i)} - \linpart \bigl(h(\vq^{f(i)}) + \rvXi\bigr) \\
	&\phantom{\;=_{\text{a.s.}}\;\dfrac{1}{N_T} \sum_{i=1,\dots, N_T}
	\mathbb{E} []}-\nlinpart\bigl(h(\vq^{f(i)} ) + \rvXi; \weight\bigr)\bigr \rVert ^2\Bigr]\\
    &=\;\dfrac{1}{N_T} \sum_{i=1,\dots, N_T}   \expectation{A(\rvQ^f, \rvXi; \weight)| \rvQ^f = \vq^{f(i)}},
\end{aligned}
\end{equation}
$\probability-$almost surely, where
\begin{equation}
A(\vq, \vek{\xi}; \weight) \equiv \bigl \lVert \vq - \linpart \bigl(h(\vq) + \vek{\xi} \bigr) - \nlinpart\bigl(h(\vq) + \vek{\xi}; \weight\bigr)\bigr \rVert^2.
\end{equation}
Let $\widehat{\mse}^{\textnormal{vr}*}(\weight | \mathcal{D}_T)$ denote the right-hand-side term in Eq.~(\ref{eq:MC_augmenteddata}).
To gain some further insight, let us assume that the ensemble members ${q^{f(i)}}, \; i =1, 2, \dots , N$, are i.i.d. samples for the sake of simplicity.
We can then approximately quantify the statistical errors of the approximation
$\widehat{\mse}(\weight | \mathcal{D}_T)$
and $\widehat{\mse}^{\textnormal{vr}}(\weight | \mathcal{D}_T)$, respectively, as
\begin{subequations}
	\begin{align}
&\variance{\widehat{\mse}(\weight | \mathcal{D}_T)- \mse(\weight)} = \dfrac{ \variance{A(\rvQ^f, \rvXi; \weight) }}{N_T}, \\
&\variance{\widehat{\mse}^{\textnormal{vr}*}(\weight | \mathcal{D}_T^a)- \mse(\weight)}= \dfrac{ \variance{ \expectation{A(\rvQ^f, \rvXi; \weight)| \rvQ^f}}}{N_T}.
	\end{align}
\end{subequations}
Using the law of total variance,
\begin{equation}
\variance{ A(\rvQ^f, \rvXi; \weight)} = \variance{ \expectation{A(\rvQ^f, \rvXi; \weight)| \rvQ^f}} + \expectation{\variance{A(\rvQ^f, \rvXi; \weight)| \rvQ^f}},
\end{equation}
we obtain
\begin{equation}
\variance{ \expectation{A(\rvQ^f, \rvXi; \weight)| \rvQ^f}}
\leq  \variance{ A(\rvQ^f, \rvXi; \weight)}.
\end{equation}

%% file: macros_v2.tex
%notations: 
%### FOR DEEP-LEARNING FILTER
\newcommand{\yob}{\mathrm{y}^{\mathrm{obs}}}
\newcommand{\mse}{J}
\newcommand{\dmodel}{\varPsi}
\newcommand{\linpart}{g_{\linear}}
\newcommand{\nlinpart}{g_{\mathrm{NN}}}
\newcommand{\linear}{\mathrm{l}}
\newcommand{\analysis} {updated}
\newcommand{\weight}{\bm{\theta}}
\newcommand{\sYob}{\const{Y}^{\mathrm{obs}}}
\newcommand{\vqtrue}{\cvq^{\mathrm{tr}}}
\newcommand{\qtrue}{\cvq^{\mathrm{tr}}}
\newcommand{\itn}{{i=1}^N}

%#### SCALAR, VECTOR, MATRIX, TENSOR
%scalar constant \mathrm: serifs, non-bold, straight
\newcommand{\const}[1]{\mathrm{#1}}
\newcommand{\cq}{\const{q}}
\newcommand{\cvq}{\const{\vq}}
\newcommand{\cmI}{\mathbf{I}}
\newcommand{\cvzero}{\mathrm{0}}
%scalar:  serifs, non-bold
%vector: serifs, non-bold
\newcommand{\vek}[1]{#1}
\newcommand{\valpha}{\bm{\alpha}}
\newcommand{\va}{\vek{a}}
\newcommand{\vq}{\vek{q}}
\newcommand{\vx}{\vek{x}}
\newcommand{\vxi}{\vek{\xi}}
\newcommand{\vy}{\vek{y}}
\newcommand{\vzeros}{\vek{0}}

%matrix: serifs, bold, upper case\\
\newcommand{\mat}[1]{\bm{#1}}
\newcommand{\mA}{\mat{A}}
\newcommand{\ma}{\mat{a}}
\newcommand{\mQ}{\mat{Q}}
\newcommand{\mq}{\mat{q}}
\newcommand{\mX}{\mat{X}}
\newcommand{\mx}{\mat{x}}
\newcommand{\mXi}{\mat{\varXi}}
\newcommand{\mxi}{\mat{\xi}}
\newcommand{\mY}{\mat{Y}}
\newcommand{\my}{\mat{y}}
\newcommand{\mZ}{\mat{Z}}
\newcommand{\mz}{\mat{z}}
%tensor:  sans-serifs, bold, italic, $\textsf{\textbf{\textsl{q}}}= q_i^{jk}, \textsf{\textbf{\textsl{f}}}= q_i^{jk}$
\newcommand{\tns}[1]{\textsf{\textbf{\textsl{#1}}}}
\newcommand{\tnsA}{\tns{A}}
\newcommand{\tnsQ}{\tns{Q}}
\newcommand{\tnsX}{\tns{X}}
\newcommand{\tnsXi}{\tns{\varXi}}
\newcommand{\tnsY}{\tns{Y}}
%random variable, vector, matrix: serifs, non-bold, capital letter\\
\newcommand{\rvek}[1]{#1}
\newcommand{\rvQ}{Q}
\newcommand{\rvX}{X}
\newcommand{\rvXi}{\varXi}
\newcommand{\rvY}{Y}
\newcommand{\rvW}{W}
% operators
\newcommand{\secondnorm}[1]{||#1||}
\newcommand{\tr}{\textnormal{tr}}%trace

%random tensor & $\textsf{\textbf{\textsl{Q}}}$ & sans-serifs, non-bold, capital letter\\

%#### PROBABILITY
% probability space
\newcommand{\sigQ}{{\mathfrak{B}}} % algebra generating by \setclosedsubsets
\newcommand{\eventalgebra}{\mathfrak{A}}
\newcommand{\eventSubalgebra}{\mathfrak{G}}
\newcommand{\eventset}{\varOmega}
\newcommand{\probability}{\mathbb{P}}
\newcommand{\probspace}{(\eventset, \eventalgebra,\probability )}
\newcommand{\subprobspace}{(\eventset, \eventSubalgebra,\probability )}
\newcommand{\measurableset}{B}
\newcommand{\expectation}[1]{\mathbb{E}\left[ #1\right]}
\newcommand{\variance}[1]{\mathbb{V}\textnormal{ar}\left[ #1\right]}
\newcommand{\cov}[1]{\mathbb{C}\textnormal{ov}\left[ #1\right]}
% conditional expectations
\newcommand{\ceX}{\varPhi_{\rvX|\rvY}}
\newcommand{\cePsiX}{\varPhi_{\varPsi(\rvX)|\rvY}}
\newcommand{\almeasurement}{\mathcal{S}_{\rvY}}
\newcommand{\spaceCE}{L_2(\mathbb{R}^m, \mathbb{R}^n, \probability_{\rvY^f})}

%#### SETS
\newcommand{\sC}{\mathbb{C}}
\newcommand{\sR}{\mathbb{R}}
\newcommand{\sZ}{\mathbb{Z}}
\newcommand{\sN}{\mathbb{N}}
%#### TERMS
\newcommand{\pdf}{{\textnormal{density}}}
\newcommand{\df}{{\textnormal{DF}}}
\newcommand{\rv}{\textnormal{RV}}
\newcommand{\pce}{{\textnormal{gPCE}}}
\newcommand{\cdf}{{\textnormal{CDF}}}

%### Operator

\newcommand{\dint}{\textnormal{d}}
\newcommand{\trans}{\intercal}